%% file: main.tex
\documentclass[preprint,12pt]{elsarticle}
\usepackage[a4paper,margin=2.5cm]{geometry}

\usepackage[utf8]{inputenc}
\usepackage{amssymb}
\usepackage{amsmath}
\usepackage{amsthm}

\usepackage{graphicx}
\usepackage{multirow}
\usepackage{subcaption}
\usepackage{amsmath,amssymb,amsfonts}
\usepackage{mathtools}
\usepackage{mathrsfs}
\usepackage[title]{appendix}
\usepackage{xcolor}
\usepackage{colortbl}
\usepackage{textcomp}
\usepackage{manyfoot}
\usepackage{booktabs}
\usepackage{algorithm}
\usepackage{algorithmicx}
\usepackage{algpseudocode}
\usepackage{listings}
\usepackage{pifont}
\usepackage{multirow}
\usepackage[normalem]{ulem}
\usepackage{comment}
\definecolor{mistyrose}{rgb}{1.0, 0.89, 0.88}
\definecolor{headerblue}{RGB}{31, 73, 125}
\definecolor{baselineyellow}{RGB}{255, 243, 176}
\definecolor{bestgreen}{RGB}{198, 239, 206}
\definecolor{headertext}{RGB}{255, 255, 255}
\usepackage{hyperref}

\hypersetup{
    colorlinks=true,       
    linkcolor=blue,        
    citecolor=blue,        
    urlcolor=blue,         
    pdfborder={0 0 0}      
}

\begin{document}

\begin{frontmatter}
\title{Evaluation of Optimisation and Bayesian Inference Methods for Reaction Rates in Atmospheric Chemical Mechanisms}

\author[lut,amc]{Valery Ashu}
\author[amc,inar]{Wenqing Peng}
\author[lut,amc]{Zhi-Song Liu}
\author[lut,amc]{Heikki Haario}
\author[saarland]{Andreas Rupp}
\author[lut,amc]{Taiwo Ashu}
\author[amc,inar]{Petri Clusius}
\author[pinumerics,inar]{Lukas Pichelstorfer}
\author[amc,inar]{Zihao Fu}
\author[lut,amc,inar]{Michael Boy}

\affiliation[lut]{
    organization={Department of Computational Engineering, LUT University},
    addressline={},
    postcode={15210},
    city={Lahti},
    country={Finland}
}

\affiliation[saarland]{
    organization={Saarland University},
    addressline={Campus E1.1},
    postcode={66123},
    city={Saarbrücken},
    country={Germany}
}

\affiliation[pinumerics]{
    organization={pi-numerics},
    city={Neumarkt am W.},
    postcode={5202},
    country={Austria}
}

\affiliation[amc]{
    organization={Atmospheric Modelling Center-Lahti},
    addressline={15140},
    postcode={15140},
    city={Lahti},
    country={Finland}
}

\affiliation[inar]{
    organization={Institute for Atmospheric and Earth System Research / Physics, Faculty of Science, University of Helsinki},
    addressline={00014},
    city={Helsinki},
    country={Finland}
}

\begin{abstract}
Constraining reaction rate coefficients is a central challenge in the development of explicit atmospheric chemical mechanisms, particularly for autoxidation systems where many reaction pathways are only indirectly observed through high-resolution mass spectrometry. In this study, we evaluate rate-coefficient optimisation methods for a toy-case autoxidation mechanism using synthetic data with known ground truth. Two complementary approaches are compared: ODE-constrained neural-network optimisation, which provides efficient point estimates of uncertain rate coefficients, and the Markov Chain Monte Carlo (MCMC) approach, which samples the posterior distribution of rate coefficients and quantifies parameter uncertainty. The methods are tested using direct concentration observations and mass-spectral observations under different noise levels. For unperturbed and low-noise synthetic observations, both methods converged towards the known rate coefficients, with the neural-network optimiser providing faster point estimates. Under high-noise conditions 
(with the signal-to-noise ratio approximately S/N${}=1$), however, MCMC was substantially more robust in recovering the rate coefficients.
The posterior analysis shows that mass-spectral aggregation broadens credible intervals even at low noise, and that high-noise mass spectra can leave many individual reaction rates weakly identifiable. Posterior predictive validation nevertheless shows how broad parameter uncertainty constrained by MCMC remains consistent with accurate reproduction of the observable mass spectrum. These results demonstrate that point-estimation and Bayesian sampling methods provide complementary information: neural-network optimisation is effective for informative data, whereas MCMC is essential for diagnosing uncertainty, non-uniqueness, and identifiability in noisy or aggregated inverse problems.
\end{abstract}


\end{frontmatter}

\input{Section/01_Introduction}

\input{Section/02_Methods}

\input{Section/02_1_autoAPRAMfw}

\input{Section/02_7_Design_of_experiment}

\input{Section/02_8_Dimensionality_Reduction}

\input{Section/02_4_NN}

\input{Section/02_5_MCMC}



\input{Section/03_Results}

\input{Section/04_Conclusion}

\input{Section/0_1_Environmental_Significance}

\input{Section/0_2_Data_Availability}

\input{Section/0_3_Author_Contribution}

\input{Section/acknowledgements}

\bibliographystyle{elsarticle-harv} 
\bibliography{main, nn}

\end{document}

%% file: Section/01_Introduction.tex
\section{Introduction}

Volatile organic compounds (VOCs) are emitted in vast quantities from both natural and anthropogenic sources, including vegetation, fossil-fuel combustion, industrial processes, and residential heating \citep{Guenther2012MEGAN21,kanakaidou}. Once released into the atmosphere, VOCs undergo complex oxidation cascades initiated by hydroxyl radicals (OH), ozone ($\mathrm{O_3}$), nitrate radicals ($\mathrm{NO_3}$), and other atmospheric oxidants, producing a chemically diverse array of oxidation products \citep{Seinfeld2016AtmosphericChemistry,Atkinson2003VOCDegradation}. Among the most significant of these processes is autoxidation, in which peroxy radicals ($\mathrm{RO_2}$) undergo rapid intramolecular hydrogen shifts followed by sequential $\mathrm{O_2}$ additions, generating highly oxygenated organic molecules (HOMs) without requiring additional external oxidants \citep{Ehn2014LowVolatilitySOA,Bianchi2019HOM}. HOMs are characterised by extremely low volatility and readily condense onto pre-existing particles, making them important contributors to secondary organic aerosol (SOA) formation and particle growth \citep{Roldin2019HOMBoreal,Bianchi2019HOM}. SOA constitutes a major fraction of submicron aerosol mass in many environments and has substantial, yet still poorly understood, effects on climate forcing, cloud condensation nuclei (CCN) concentrations, and human health through its contribution to fine particulate matter ($\mathrm{PM_{2.5}}$) \citep{Jimenez2009OrganicAerosols,Hallquist2009SOA}.

Despite decades of research, a persistent gap remains between observed and modelled SOA mass concentrations, particularly during urban haze and pollution episodes. One reason for this discrepancy is the incomplete representation of autoxidation chemistry in regional and global chemical transport models \citep{Bianchi2019HOM,McFiggans2019SOAMixtures}. Aromatic VOCs constitute a substantial fraction of urban VOC emissions from traffic, solvent use, and combustion, but their autoxidation chemistry remains especially poorly represented. The corresponding reaction networks can contain hundreds or thousands of intermediate species and reaction pathways \citep{Atkinson2003VOCDegradation,Bianchi2019HOM}. Even for well-studied systems such as $\alpha$-pinene and benzene, near-realistic autoxidation mechanisms have only recently become computationally tractable \citep{Roldin2019HOMBoreal,Pichelstoerfer}. Bridging this gap requires the construction of new chemical mechanisms together with robust and scalable methods for determining and optimising their reaction rate coefficients. However, estimating these coefficients is a severely ill-posed problem because high-resolution mass spectrometry experiments typically provide only limited observational constraints. Meeting this challenge requires methods that combine data-driven estimation with physical constraints. Related data-driven surrogate approaches have demonstrated that neural architectures can efficiently recover hidden parameters in other complex dynamical systems, including attention-enhanced CNNs for inverse parameter estimation in cellular automata \citep{ashu2026attention}.

One framework developed to address the need for explicit autoxidation mechanisms is the automated alkoxy/peroxy radical autoxidation mechanism framework (autoAPRAM-fw). Building on the Master Chemical Mechanism (MCM), autoAPRAM-fw enables the modular and automated generation of $\mathrm{RO_2}$ and $\mathrm{RO}$ radical chemistry. It comprises two main components: \textit{autoReactions}, which generates chemical mechanisms and the corresponding differential equations from predefined reaction classes, and \textit{autoSMILES}, which predicts plausible molecular structures of the resulting products using SMILES notation \citep{Pichelstoerfer,smiles}. Rather than including every available reaction class, autoAPRAM-fw allows pathways to be selected according to the chemical system being modelled. The available chemistry includes autoxidation through intramolecular H-shifts and sequential $\mathrm{O_2}$ additions, together with competing $\mathrm{RO_2}$ loss pathways involving $\mathrm{NO}$, $\mathrm{HO_2}$, and other $\mathrm{RO_2}$ radicals. These pathways can produce organic nitrates, hydroperoxides, carbonyl compounds, fragmentation products, alkoxy radicals, and ROOR accretion products. The generated reaction networks can be implemented in chemical transport and aerosol models such as ARCA \citep{arcabox}, ADCHAM, ADCHEM \citep{adcham,adchem} and SOSAA \citep{petri_sosaa}.

In this work, we evaluate and compare two approaches for estimating reaction rate coefficients in a toy-case autoxidation mechanism generated using autoAPRAM-fw \citep{Pichelstoerfer}. The toy precursor was designed to resemble a chemically related molecule for which reaction-rate information is available in the literature. Chemically reasonable reference rate coefficients were assigned using reported values for this analogous system as a basis \citep{Lukas_CONSTRAINTS}. These coefficients define the known synthetic ground truth against which the inferred values are evaluated. As the first approach  physics-informed neural networks are studied. Especially, we use the ODE framework SPIN-ODE, which combines trajectory learning with ODE-constrained optimisation to obtain point estimates of the rate coefficients \citep{Peng2025SPINODE}. The Bayesian inference approach characterises the posterior distributions of rate coefficients consistent with the  observations. For this purpose we employ the Delayed Rejection Adaptive Metropolis Markov Chain Monte Carlo (DRAM-MCMC) algorithm,  \citep{Haario2006DRAM}. These methods provide complementary information: SPIN-ODE produces efficient point estimates, whereas DRAM-MCMC quantifies parameter uncertainty and reveals identifiability and non-uniqueness in the inverse problem. By benchmarking both approaches against synthetic data with known ground truth, this study assesses their respective strengths and limitations for constraining reaction rate coefficients in autoxidation mechanisms.

%% file: Section/02_Methods.tex
\section{Methods}
In this section, we introduce the simplified “toy-case” chemistry scheme, used both as a test-bed for all methods and as the basis for generating synthetic data. We then provide the technical specifications of the methods and outline their implementation.

\subsection{Preliminaries}
Chemical kinetics determines reaction rates as functions of the concentrations of the reactant species and the corresponding rate coefficients. The $i$-th reaction in a reaction system can be generally expressed as
\begin{equation}\label{eq:reaction}
    s_{i,a}^f\textrm{A} + s_{i,b}^f\textrm{B} \leftrightarrow s_{i,a}^r\textrm{C} + s_{i,b}^r\textrm{D},
\end{equation}
where $s_{i,j}^f$ and $s_{i,j}^r$ denote the forward and reverse stoichiometric coefficients for species $j$, which represent the number of molecules consumed or produced in a single reaction event.

The reaction rate is given by the product of the rate coefficient $k_i$ and the concentrations of the reactants raised to their stoichiometric powers:
\begin{equation} \label{eq:rate}
    r_i = k_i[A]^{s_{i,a}^f}[B]^{s_{i,b}^f}.
\end{equation}
The full reaction system forms a system of ordinary differential equations (ODEs), where the concentration of each species evolves according to
\begin{equation}\label{eq:duidt}
    \frac{d[u_j]}{dt} = \sum_{i} -s_{i,j}^f r_i + s_{i,j}^r r_i,
\end{equation}
equivalently,
\begin{equation}\label{eq:dydt}
\frac{d\boldsymbol{u}}{dt} = (S^r - S^f) \,\boldsymbol{r} = S\,\boldsymbol{r}=S\,r(\boldsymbol{u}, \boldsymbol{k}),
\end{equation}
where $S^f$and $S^r$ are the forward and reverse stoichiometric matrices of $s_{i,j}^f$ and $s_{i,j}^r$, $S$ is the net stoichiometric matrix, $\boldsymbol{r}$ is the reaction rate vector of $r_i$, $\boldsymbol{u}$ is the species concentrations vector of $u_i$, and $\boldsymbol{k}$ is the rate coefficients vector of $k_i$.

We note that the right-hand side of the differential equation is linear with respect to the  reaction rate parameters $k_i$. Thus, if all the components $y_j$ could be measured at dense enough time points, the task of optimizing the reaction rates would boil down to solving a straightforward overdetermined linear equation system. However, typically only a small subset of components can be measured. Moreover,
atmospheric chemical systems are typically stiff, characterized by large disparities between slow and fast timescales.
Efficient stiff ODE solvers  employ specialized numerical techniques (e.g., implicit integration) to ensure stability and accuracy.

\subsection{Observability and derivative information}
\label{sec:prelim_interp}

The preceding remark can be made precise, and doing so identifies the three effects that
make rate-coefficient estimation ill-posed in practice: the observation operator removes
part of the state, the required time derivatives are not measured, and the stiffness of the
system separates the timescales on which those derivatives carry information.

\subsubsection{Linearity in the rate coefficients}

Collecting the reactant monomials of equation~\eqref{eq:rate} into the vector
$\boldsymbol{\varphi}(\boldsymbol{u})\in\mathbb{R}^{n_r}$ with
$\varphi_i(\boldsymbol{u}) = \prod_{j} u_j^{\,s_{i,j}^f}$, the rate vector factorises as
$\boldsymbol{r}(\boldsymbol{u},\boldsymbol{k}) = \operatorname{diag}(\boldsymbol{k})\,\boldsymbol{\varphi}(\boldsymbol{u})$
and equation~\eqref{eq:dydt} becomes
\begin{equation}\label{eq:linear_k}
    \frac{d\boldsymbol{u}}{dt} = R(\boldsymbol{u})\,\boldsymbol{k},
    \qquad
    R(\boldsymbol{u}) = S \operatorname{diag}\!\big(\boldsymbol{\varphi}(\boldsymbol{u})\big)
    \in \mathbb{R}^{n_s \times n_r},
\end{equation}
with $n_s$ species and $n_r$ reactions. The regressor matrix $R(\boldsymbol{u})$ depends on
the state but not on the unknown coefficients. If the state and its derivative were known at
time levels $t_1,\dots,t_Q$, the coefficients would follow from the weighted linear
least-squares problem
\begin{equation}\label{eq:normal_eq}
    \hat{\boldsymbol{k}} = \arg\min_{\boldsymbol{k}} \sum_{j=1}^{Q}
      \big\| W\big(\dot{\boldsymbol{u}}(t_j) - R(\boldsymbol{u}(t_j))\,\boldsymbol{k}\big) \big\|_2^2,
    \qquad G\,\hat{\boldsymbol{k}} = \boldsymbol{b},
\end{equation}
with Gram matrix $G=\sum_j R^\top W^\top W R$ and $W$ a diagonal weighting that compensates
for concentrations differing by orders of magnitude. The coefficients are identifiable
from~\eqref{eq:normal_eq} if and only if $G$ is non-singular, that is, if the regressors
sampled along the trajectories span $\mathbb{R}^{n_r}$.

\subsubsection{Partial and aggregated observations}

Equation~\eqref{eq:normal_eq} assumes access to the full state. The available data are
instead noisy samples at $P$ sparse time levels,
\begin{equation}\label{eq:obs_model}
    \boldsymbol{y}_p = H\,\boldsymbol{u}(t_p) + \boldsymbol{e}_p,
    \qquad p = 1,\dots,P,
\end{equation}
where $H\in\mathbb{R}^{n_m\times n_s}$ is the observation operator and $\boldsymbol{e}_p$ the
observation error. Two cases are relevant here. When individual species concentrations are
recorded, $H$ selects the observed rows and $n_m$ is the number of measured species. When
the observable is a mass spectrum, $H$ is a $0/1$ matrix that sums all species sharing the
same nominal molecular mass, so $n_m$ is the number of populated mass channels and each row
of $H$ merges several species.

Only the projected dynamics are then accessible,
\begin{equation}\label{eq:proj_dynamics}
    H\frac{d\boldsymbol{u}}{dt} = H R(\boldsymbol{u})\,\boldsymbol{k},
\end{equation}
so the effective regressor is $HR(\boldsymbol{u})$, whose rank is at most
$\min(n_m,n_r)$. Whenever $n_m<n_r$, the projected system is rank-deficient at every time
level, and any $\boldsymbol{\delta}\in\mathcal{N}\big(HR(\boldsymbol{u})\big)$ leaves the
observations unchanged to first order: entire directions in coefficient space are invisible.
Aggregation is the more severe case, because merging species into a common channel both
reduces $n_m$ and mixes rows of $R$ that would otherwise constrain different reactions.
Sampling additional time levels and additional experiments enlarges the union of the row
spaces and can restore formal rank, but the resulting $G$ typically remains
ill-conditioned, so the least-squares solution is highly sensitive to perturbations in
$\boldsymbol{b}$. A further consequence of~\eqref{eq:obs_model} is that
$R(\boldsymbol{u}(t_j))$ itself is not directly evaluable, since the unobserved species
entering the monomials $\boldsymbol{\varphi}$ must be reconstructed rather than measured.

\subsubsection{Interpolation and numerical differentiation}

The derivative $\dot{\boldsymbol{u}}$ required by~\eqref{eq:normal_eq} is never measured and
must be inferred from the samples~\eqref{eq:obs_model}. This requires first replacing the
$P$ discrete samples by a continuous curve. In general form, such a curve solves a penalised
interpolation problem over a hypothesis class $\mathcal{U}$,
\begin{equation}\label{eq:smoothing}
    \hat{\boldsymbol{u}} = \arg\min_{\boldsymbol{v}\in\mathcal{U}}
    \sum_{p=1}^{P} \big\| H\boldsymbol{v}(t_p) - \boldsymbol{y}_p \big\|_W^2
    + \lambda\, \mathcal{R}(\boldsymbol{v}),
\end{equation}
where $\mathcal{R}$ is a roughness functional, for example
$\mathcal{R}(\boldsymbol{v})=\int\|\ddot{\boldsymbol{v}}(t)\|^2dt$, and $\lambda\ge0$
balances data fidelity against smoothness. Evaluating $\hat{\boldsymbol{u}}$ within the
sampled interval $[t_1,t_P]$ is \emph{interpolation}; evaluating it outside that interval,
or from initial conditions not represented in the data, is \emph{extrapolation}, for which
no accuracy guarantee follows from~\eqref{eq:smoothing}.

Differentiating the interpolant is ill-posed. For a second-order central difference with
step $h$ and a perturbation of magnitude $\delta$ in the differentiated curve,
\begin{equation}\label{eq:fd_error}
    \Big\| \dot{\boldsymbol{u}}(t) - \tfrac{\hat{\boldsymbol{u}}(t+h)-\hat{\boldsymbol{u}}(t-h)}{2h} \Big\|
    \le \tfrac{h^2}{6}\big\|\dddot{\boldsymbol{u}}\big\|_\infty + \tfrac{\delta}{h},
\end{equation}
which is minimised at $h^\star=\big(3\delta/\|\dddot{\boldsymbol{u}}\|_\infty\big)^{1/3}$
and leaves an irreducible error of order $\delta^{2/3}$. Refining the temporal sampling alone
therefore cannot remove the derivative error; only a reduction of $\delta$, obtained through
the smoothing in~\eqref{eq:smoothing}, makes small $h$ admissible.

\subsubsection{Smoothing bias under stiffness}

Smoothing in turn interacts with the stiffness noted above. Linearising
equation~\eqref{eq:dydt} about a reference trajectory gives the Jacobian
$J=\partial(S\boldsymbol{r})/\partial\boldsymbol{u}$ with eigenvalues $\lambda_m$ and
stiffness ratio
\begin{equation}\label{eq:stiffness}
    \kappa = \frac{\max_m |\mathrm{Re}\,\lambda_m|}{\min_m |\mathrm{Re}\,\lambda_m|},
\end{equation}
which is large for atmospheric radical systems. Any interpolant of the
form~\eqref{eq:smoothing} acts as a low-pass filter with an effective cut-off $\omega_c$ set
by $\lambda$ and by the sampling density. Modes with $|\lambda_m|\gg\omega_c$ are attenuated
in $\hat{\boldsymbol{u}}$ and are therefore absent from the finite-difference derivatives, so
the coefficients of the corresponding fast reactions are systematically biased towards
smaller values, irrespective of the amount of data.

Equations~\eqref{eq:proj_dynamics}, \eqref{eq:fd_error} and~\eqref{eq:stiffness} thus
summarise the three obstacles that any estimate of $\boldsymbol{k}$ must confront: a
rank-deficient observation operator, noise amplification in differentiation, and attenuation
of the fast dynamics that carry information on the most reactive channels.

%% file: Section/02_1_autoAPRAMfw.tex
\subsection{Toy-case chemistry scheme and synthetic data}
The toy-case autoxidation mechanism was generated using autoAPRAM-fw \citep{Pichelstoerfer} for a model benzene derivative with the molecular formula $\mathrm{C_7H_8}$. The resulting mechanism comprised 45 active chemical species and 50 reactions. It included VOC oxidation initiated by OH and $\mathrm{O_3}$, $\mathrm{RO_2}$ autoxidation through intramolecular H-shifts and subsequent $\mathrm{O_2}$ addition, and competing $\mathrm{RO_2}$ reactions with NO, $\mathrm{HO_2}$, and other $\mathrm{RO_2}$ radicals. These pathways produced organic nitrates, hydroperoxides, alcohols, carbonyl compounds, alkoxy radicals, fragmentation products, and ROOR accretion products. Chemically reasonable reference rate coefficients and branching fractions were assigned using published information for analogous molecules and reactions as a basis \citep{Lukas_CONSTRAINTS}. These values defined the known synthetic ground truth against which the rate coefficients inferred by SPIN-ODE and DRAM-MCMC were evaluated. The complete mechanism, including all chemical species, reactions, rate expressions, and branching fractions, is provided in Supplementary Table~S1.

%% file: Section/02_7_Design_of_experiment.tex
\subsection{Design of Experiments (DOE)}
A set of ten simulation experiments was designed to investigate the sensitivity of the model results to variations in key initial chemical conditions. Four input parameters were varied across the experiment set: the initial concentrations of NO, OH, and HO$_2$, and the benzene derivative (toy-VOC). All other environmental and background conditions were held constant for all runs, including temperature (288~K), pressure (1000~hPa), ozone concentration ($5 \times 10^{11}$~cm$^{-3}$), and the sinusoidal nitrate constraint, so that differences in the model response could be attributed solely to the selected variables.

The experiments were constructed to span a wide range of chemically relevant conditions by sampling combinations of low and high values for each factor, together with a small number of intermediate levels for selected parameters. Initial NO concentrations ranged from $2 \times 10^{7}\,\mathrm{cm}^{-3}$ to $2 \times 10^{11}$~cm$^{-3}$, with an additional mid-range case at $2 \times 10^{9}$~cm$^{-3}$. Initial OH concentrations were given at $1 \times 10^{6}$ and $1 \times 10^{8}$~cm$^{-3}$. The TOY parameter was sampled between $1 \times 10^{11}$ and $1 \times 10^{14}$~cm$^{-3}$, with intermediate cases at $5 \times 10^{12}$~cm$^{-3}$. Initial HO$_2$ concentrations ranged from $1 \times 10^{8}$ to $5 \times 10^{9}$~cm$^{-3}$, with one additional intermediate case at $1 \times 10^{9}$~cm$^{-3}$. The complete set of ten input combinations is provided in Supplementary Table~S3.

This design represents a screening-type experiment intended to explore model behaviour across the multidimensional concentration space and to reveal the dominant sensitivities of different reaction rates to the selected inputs. Because the mechanism contains many rate parameters with poorly constrained initial estimates, a broad screening design was considered more appropriate than formal optimal-design approaches, such as D-optimality, which generally require prior parameter estimates for nonlinear models. We note that only such an ensemble of experiments was able to identify the reaction rate parameters. Any single experiment, even with highly accurate data, produced strong correlations between parameters, resulting in high uncertainties. 

\subsubsection{Construction of synthetic observations}
\label{sec:synthetic_observations}

Two observation representations were constructed from the simulated concentration trajectories. In the direct-concentration representation, the concentration of each selected chemical species was retained separately at every sampled time point. In the synthetic mass-spectral representation, each species was assigned to a nominal-mass channel, and the concentrations of all species with the same nominal molecular mass were summed:
\[
I_m(t)=\sum_{j:\,m_j=m} C_j(t),
\]
where $C_j(t)$ is the simulated concentration of species $j$, $m_j$ is its nominal molecular mass, and $I_m(t)$ is the resulting intensity in nominal-mass channel $m$. We refer to this operation as species-to-nominal-mass aggregation. It represents the loss of species-level information when multiple compounds contribute to the same nominal-mass channel; it does not represent instrumental peak broadening, overlap caused by finite instrumental resolution, or temporal averaging. The complete mapping between model species and nominal-mass channels is provided in Supplementary Table~S2.

For each experiment, ARCA was run for 5~s with a temporal resolution of 0.1~s, giving 50 simulated time levels. To mimic sparse measurements, five time points were selected from each simulation. The resulting direct-concentration observation array has dimensions $\mathbb{R}^{10\times5\times45}$, corresponding to ten experiments, five recorded time points, and 45 chemical species. The mass-spectral observations were constructed from the same concentration arrays using the species-to-nominal-mass aggregation defined above.

\paragraph{Observation-noise regimes}
The same three noise regimes were applied to both observation representations and were used to evaluate both inference methods. In the noise-free regime, the synthetic observations were
not perturbed:
\[
Y_{\mathrm{obs}}=Y_{\mathrm{clean}}.
\]

The low-noise observations were generated using a Gaussian relative-error model:
\[
  Y_{\mathrm{obs}}
  =
  Y_{\mathrm{clean}}(1+\sigma R),
  \qquad
  R\sim\mathcal{N}(0,1),
\]
  with $\sigma=0.01$. Equivalently,
\[
  Y_{\mathrm{obs}}
  =
  Y_{\mathrm{clean}}
  +
  \sigma Y_{\mathrm{clean}}R,
\]
so the perturbation is additive but heteroscedastic on the original scale, with a standard deviation proportional to the clean signal. The corresponding relative error has a standard
deviation of 1\,\%.

The high-noise observations were generated using a lognormal relative-error model:
\[
  Y_{\mathrm{obs}}
  =
  Y_{\mathrm{clean}}\exp(\sigma R),
  \qquad
  R\sim\mathcal{N}(0,1),
\]
with $\sigma=\log_{10}(2)$. Thus, one standard deviation in logarithmic space corresponds to a factor-of-two change on the original concentration or intensity scale. The perturbation is
multiplicative on the original scale but additive in logarithmic space:
\[
  \log(Y_{\mathrm{obs}})
  =
  \log(Y_{\mathrm{clean}})
  +
  \sigma R.
\]

%% file: Section/02_8_Dimensionality_Reduction.tex
\subsection{Dimensionality Reduction}

The dimensionality reduction used here follows the reaction-invariant and reaction-variant decomposition introduced for reactive transport systems by Knabner, Kräutle, Hoffmann and co-workers \cite{hoffmann2010monograph,hoffmann2010parallel,hoffmann2012general,kraeutle2005new,kraeutle2007reduction}. It allows replacing the ODE system by a smaller and less stiff ODE system that yields the same result as the unreduced system. Thus, it can significantly boost computational performance, particularly in reactive systems with many chemical species but and comparably few independent reactions. Writing the kinetic system in stoichiometric form as
\[
 \frac{\mathrm d \boldsymbol u}{\mathrm dt} = \hat S \, \hat{\boldsymbol r}(\boldsymbol u),
\]
where \( \boldsymbol u \in \mathbb{R}^{n_s} \) contains the concentrations of chemical species, \( \hat S \in \mathbb{R}^{n_s \times n_r} \) is the stoichiometric matrix, and \( \hat{\boldsymbol r} \) is the vector of elementary reaction rates. The complete \(45\times50\) stoichiometric matrix used in this study is provided in Supplementary Figure~S6. The idea is to replace the original species variables by reaction variants and reaction invariants. If the stoichiometric columns are linearly independent, a matrix \( U \) is chosen whose columns span the orthogonal complement of the space spanned by the columns of \( S \) and the matrix is written.
\[
 \boldsymbol u = \hat S \boldsymbol \xi + U \boldsymbol \eta, \qquad \boldsymbol \xi = \hat S^\star \boldsymbol u, \qquad \boldsymbol \eta = U^\star \boldsymbol u,
\]
with \( \hat S^\star = (\hat S^\top \hat S)^{-1}\hat S^\top \) and, analogously, \( U^\star = (U^\top U)^{-1}U^\top \). 
The variables \( \boldsymbol \xi \) evolve only in the directions affected by reactions, whereas \( \boldsymbol \eta \) represents conserved combinations of species. However, this construction cannot be applied to the present toy problem straight out of the box: since the mechanism contains \( 45 \) active chemical species and \( 50 \) reactions, the stoichiometric matrix does not consist only of linearly independent columns. Therefore, one first has to decompose
\[
 \hat S = S \widetilde S,
\]
where \( S \in \mathbb{R}^{n_s \times q} \) has full column rank \( q=\operatorname{rank}(\hat S) \) and \( \widetilde S \in \mathbb{R}^{q \times n_r} \) expresses every original reaction vector as a linear combination of the independent columns of \( S \). In practice, \( S \) may be obtained by selecting a maximal independent subset of columns of \( \hat S \), for example using a rank-revealing QR decomposition, and then setting \( \widetilde S = S^\star \hat S \) with \( S^\star=(S^\top S)^{-1}S^\top \). The original system can then be rewritten as
\[
 \frac{\mathrm d \boldsymbol u}{\mathrm dt} = S \boldsymbol r(\boldsymbol u), \qquad \boldsymbol r(\boldsymbol u) = \widetilde S \hat{\boldsymbol r}(\boldsymbol u),
\]
so that the reduction is applied to the full-rank matrix \( S \) rather than to \( \hat S \) itself. For the present toy mechanism this only yields a moderate speedup of approximately \( 10\% \), because the number of independent reaction directions remains close to the number of chemical species. In larger real-world autoxidation mechanisms, by contrast, many more elementary reactions may collapse onto a smaller number of independent stoichiometric directions, so the gap between \( n_s \) and \( q \) is expected to be larger and the same reduction scheme should lead to a more substantial computational gain.

The same framework can also accommodate equilibrium reactions, i.e. reactions that are sufficiently fast to be treated as algebraic equilibrium constraints rather than as explicit kinetic ODE terms \cite{hoffmann2012general,kraeutle2007reduction}. In this case, the reduction separates kinetic reaction directions from equilibrium constraints, so that the very fast processes no longer have to be resolved dynamically by the time integrator. For atmospheric mechanisms, this is potentially useful because fast radical interconversion or quasi-equilibrated reaction classes may otherwise dominate the stiffness of the system. Treating such reactions in equilibrium form can therefore reduce both the number of dynamic variables and the stiffness of the remaining ODE system, which in turn may permit larger stable time steps and further improve computational efficiency.

%% file: Section/02_4_NN.tex
\subsection{Neural ODE emulation and ODE-constrained optimisation}
\label{sec:nn}

The data-driven approach was first introduced in our earlier work~\cite{peng2025spin}, where reaction rate coefficients are estimated directly from time-resolved concentration trajectories, given the reaction pathways but without any prior knowledge of the coefficient values. That method proceeds in three steps. The first two initialise an estimate of the rate coefficients from arbitrary initial values: a black-box Neural ODE is trained to emulate the concentration trajectories as a smooth surrogate, and the kinetic rate coefficients are then optimised against the time derivatives of this surrogate. This derivative-matching stage is necessary because ODE-constrained optimisation cannot be applied to randomly initialised coefficients, for which the stiff time integration would fail. The third step refines the coefficients through ODE-constrained optimisation against the original concentration trajectories, recovering the fast dynamics that derivative matching on the smoothed surrogate cannot capture. Since the present task is to optimise rate coefficients that are already reasonably constrained, only the third step of the original method is applied here. The ODE-constrained optimisation was evaluated using the two observation representations and three noise regimes defined in Section~\ref{sec:synthetic_observations}. The three steps are summarised briefly below.

First, a black-box Neural ODE network is trained to fit the discrete concentration samples $\boldsymbol{u}(t)$ and to construct a continuous trajectory emulator. A multilayer perceptron (MLP) parametrised by $\theta$ learns the rate of change of each species, which is then integrated over time to predict the concentration time series $\hat{\boldsymbol{u}}(t_i)$. A gradient-descent optimiser trains the network by minimising the concentration mismatch. Once trained, the Neural ODE emulates the species concentration trajectories $\hat{\boldsymbol{u}}(t)$.
\begin{gather}
    \hat{\boldsymbol{u}}(t_i) = \boldsymbol{u}(t_0) + \int_{t_0}^{t_i}\text{MLP}_{\theta}(\boldsymbol{u}(\tau)) d\tau \qquad t_i \in \{t_1, \dots, t_P\} \\
\hat{\theta} = \arg\min_{\theta} \text{MSE}(\frac{\hat{\boldsymbol{u}}(t)}{\boldsymbol{u}(t)}, \boldsymbol{1}),
\end{gather}
where the relative error balances the contributions of species whose concentrations differ by orders of magnitude. This is the penalised interpolation problem~\eqref{eq:smoothing} with the hypothesis class $\mathcal{U}$ restricted to flows of the autonomous vector field $\text{MLP}_\theta$ and with the weighting $W$ realised as the relative-error scaling. Smoothness is imposed structurally, through the regularity of $\text{MLP}_\theta$, rather than through an explicit roughness functional, so no penalty term is required and $\lambda=0$. The mean squared error (MSE) between two vectors of size $N$ is defined as:
\begin{equation}\label{eq:MSE}
    \text{MSE}( \boldsymbol{u}, \hat{\boldsymbol{u}}) = \frac{1}{N} \sum_{i=1}^{N} \left( \hat{u}_i  - u_i  \right)^2.
\end{equation}

Second, we construct a process-based kinetic model that implements the reaction rate law of equation~\ref{eq:dydt}, with rate coefficients randomly initialised as $\hat{\boldsymbol{k}}$. We generate dense concentration trajectories $ \hat{\boldsymbol{u}}(t_j)$ using the neural emulator trained in the first step, and apply the second-order finite difference method to compute the time derivative $\dot{\boldsymbol{u}}(t_j)$. The rate coefficients are then updated by the gradient descent optimiser.
\begin{gather}
    \hat{\boldsymbol{u}}(t_j) = \boldsymbol{u}(t_0) + \int_{t_0}^{t_j}\text{MLP}_{\theta}(\boldsymbol{u}(\tau)) d\tau \qquad t_j \in \{t_1, \dots, t_Q\}, Q > P, \\
    \dot{\boldsymbol{u}}(t_j) = \frac{\hat{\boldsymbol{u}}(t_{j+1}) - \hat{\boldsymbol{u}}(t_{j-1})}{t_{j+1}-t_{j-1}}, \\
    \hat{\dot{\boldsymbol{u}}}(t) = S\,{r}(\boldsymbol{u}, \hat{\boldsymbol{k}}),\\
    \hat{\boldsymbol{k}} = \arg\min_{\hat{\boldsymbol{k}}} \text{MSE}(\frac{\hat{\dot{\boldsymbol{u}}}(t)}{\dot{\boldsymbol{u}}(t)}, \boldsymbol{1}).\label{eq:derivmatch}
\end{gather}
Because $S\,r(\boldsymbol{u},\boldsymbol{k}) = R(\boldsymbol{u})\,\boldsymbol{k}$ by equation~\eqref{eq:linear_k}, the objective~\eqref{eq:derivmatch} is the least-squares problem~\eqref{eq:normal_eq} with the weighting $W=\operatorname{diag}(\dot{\boldsymbol{u}}(t_j))^{-1}$, evaluated at the $Q$ dense time levels supplied by the emulator. Sampling the interpolant rather than the raw observations reduces the perturbation $\delta$ entering the error bound~\eqref{eq:fd_error} and therefore admits a smaller differencing step $h$.

The stiff reaction system exhibits fast and slow dynamic simultaneously. The finite differences computed from the emulated concentration trajectories neglect the fast dynamics. To recover the full dynamics, the third step integrates the time derivatives produced by the kinetic model and again uses gradient descent to optimise the rate coefficients, this time against the original concentration trajectories. The third optimisation step used in the work can be formulated as:
\begin{gather}
    \hat{\boldsymbol{u}}(t) = \boldsymbol{u}(t_0) + \int_{t_0}^{t} S\,{r}(\boldsymbol{u}, \hat{\boldsymbol{k}}) dt, \\
\hat{\boldsymbol{k}} = \arg\min_{\hat{\boldsymbol{k}} } \text{MSLE}(\boldsymbol{u}(t), \hat{\boldsymbol{u}}(t)).\label{eq:odeconstrained}
\end{gather}
The residual in~\eqref{eq:odeconstrained} is formed on integrated trajectories, so no numerical differentiation is performed and the fast modes are resolved by the stiff solver instead of being attenuated as described by~\eqref{eq:stiffness}. Both error sources of Section~\ref{sec:prelim_interp} are removed at the cost of an objective that is no longer linear in $\hat{\boldsymbol{k}}$ and that requires a stiff solve per evaluation; the integration is moreover stable only for coefficient vectors already within a physically plausible range, which is why the first two steps are needed when starting from arbitrary values. When the observable is aggregated rather than species-resolved, $\boldsymbol{u}(t)$ and $\hat{\boldsymbol{u}}(t)$ in~\eqref{eq:odeconstrained} are replaced by $H\boldsymbol{u}(t)$ and $H\hat{\boldsymbol{u}}(t)$ following~\eqref{eq:obs_model}, so the coefficients are constrained only through the projected dynamics~\eqref{eq:proj_dynamics}.
Since the species concentrations are positive and vary over several orders of magnitude, we adopt the mean squared logarithmic error (MSLE) as the objective function:
\begin{equation}\label{eq:MSLE}
    \text{MSLE}( \boldsymbol{u}, \hat{\boldsymbol{u}}) = \frac{1}{N} \sum_{i=1}^{N} \left( \log(\hat{u}_i + \varepsilon) - \log(u_i + \varepsilon) \right)^2,
\end{equation}
where $\varepsilon$ is a small positive constant that keeps the logarithm finite when a concentration approaches zero.

%% file: Section/02_5_MCMC.tex
\subsection{MCMC method used in this study}

The Bayesian analysis was designed to quantify the uncertainty and identifiability of reaction rate coefficients in the toy autoxidation mechanism. Rather than estimating a single optimal set of rates, the MCMC approach samples the posterior distribution of all rate multipliers that are statistically consistent with the observations. This is important for the present parameter estimation problem because different combinations of reaction rates can lead to similar concentration trajectories or mass spectra, especially when only sparse or aggregated observations are available.  A point estimate alone cannot show whether a reaction rate is tightly constrained or whether it is one of many values that fit the observations nearly equally well.

Markov Chain Monte Carlo (MCMC) is a Bayesian sampling approach used to approximate a posterior probability distribution when it cannot be evaluated analytically. In parameter-estimation problems, the posterior combines prior information about the parameters with the likelihood of the observations given a model prediction.
%
%
%
The posterior distribution was sampled using the Delayed Rejection Adaptive Metropolis (DRAM) algorithm implemented in the MCMCStat MATLAB toolbox \citep{Haario2006DRAM, laine2023mcmcstat}. DRAM was chosen because the reaction-rate inverse problem is high-dimensional and can contain strong parameter correlations. The adaptive component updates the proposal covariance using the sampled chain, while the delayed-rejection component allows a second, smaller proposal after a rejected move. This improves sampling robustness when the posterior contains narrow or correlated directions.

The MCMC analysis used the two observation representations and three noise regimes defined in Section~\ref{sec:synthetic_observations}. The concentration cases retained the selected chemical species individually, whereas the mass-spectral cases used species-to-nominal-mass aggregation.

Although the synthetic observations in the noise-free cases were not perturbed, evaluation of the MCMC likelihood required a strictly positive residual scale. An exactly zero variance would produce a degenerate likelihood and lead to division by zero in the normalised residuals. Therefore, for the noise-free MCMC cases, we used
\[
  \sigma_{\log}=10^{-6}.
\]
This value represents numerical regularisation of the likelihood rather than measurement noise added to the synthetic observations.

The likelihood was evaluated using logarithmic residuals. For an observation matrix $Y$ and a model prediction $Y_\theta$, the residual was computed as:
\[
    r(\theta)=
    \frac{
    \log\left(\max(Y,Y_{\min})\right)
    -
    \log\left(\max(Y_\theta,Y_{\min})\right)
    }{\sigma_{\log}},
\]
Here, $Y_{\min}=10^{-12}$ prevents singular logarithms and $\sigma_{\log}$ is the log-residual scale. As described above, $\sigma_{\log}=10^{-6}$ was used for the noise-free MCMC cases; for the low- and high-noise cases, the residual scale was set to the corresponding observation-noise scale. The cost function supplied to the MCMC sampler was the residual sum of squares,
\[
    \mathrm{RSS}(\theta)=\sum r(\theta)^2.
\]
summed over all selected species or mass channels, time points, and experiment cases. For the high-noise cases, this log-space residual directly matches the lognormal noise model. For the low-noise cases, it is a close approximation to the linear multiplicative perturbation because $\log(1+\sigma R)\approx\sigma R$ when $\sigma=0.01$.

For each observation case, the model output was transformed into the corresponding observation representation defined in Section~\ref{sec:synthetic_observations}. The residual was then computed in logarithmic space between the synthetic observations and the corresponding model predictions. Each chain contained $10^6$ samples, and posterior summaries were computed after discarding the first 20\,\% as burn-in.

Each inferred parameter was represented as a dimensionless multiplier applied to a nominal reaction rate coefficient. If $k_j^{0}$ is the nominal rate coefficient for reaction $j$, the rate used in the simulation was
\[
    k_j = \theta_j k_j^{0},
\]
where $\theta_j$ is the sampled multiplier. The synthetic reference data were generated with $\theta_j=1$ for all reactions. The posterior distribution of $\theta_j$ therefore indicates both the ability of the observations to recover the known synthetic truth and the remaining uncertainty in each reaction rate.
A bounded uniform prior was used for each multiplier,
\[
    \theta_j \sim \mathcal{U}(0.1,10).
\]
This prior allows each selected reaction rate to vary over two orders of magnitude while keeping the mechanism within a finite and physically meaningful parameter range. Parameter proposals outside this interval were rejected by assigning a very large cost. 

%% file: Section/03_Results.tex
\section{Results}
\subsection{Overall comparison between MCMC and Neural Network}

The noise definitions of the previous section were applied to both direct concentration and mass-spectral observations. Both MCMC and Neural Network need to have an initial estimation of the rate coefficients. Here, a Gaussian noise of 10\% in log scale is applied to the true rate coefficients to construct the initial values.
Define the root mean squared relative error (RMSRE) between two vectors $\hat{k},k$ as
\begin{equation}
    \text{RMSRE}(\hat{k}, k) = \sqrt{\frac{1}{n} \sum_{i=1}^{n} \left( 1 - \frac{\hat{k}_i}{k_i} \right)^2}
\end{equation}
We keep the same initial guesses fixed for all experiments. The RMSRE error between the true and the randomized  initial rate coefficients  assumed the value 0.0928.

Table~\ref{tab:rmsre} presents the overall RMSRE comparison between MCMC and Neural Network under different noise levels. In the noise-free case, the Neural Network optimization 
converges to the correct true parameter values, while the outcome of MCMC, selected as the median of the sampled parameter chain, depends on  the numerical value given to   $\sigma_{\log}$.
Under mild noise, both methods show modest performance degradation, yet the Neural Network remains competitive with MCMC on concentration data (0.0114 vs.\ 0.0146) and mass spectrum data (0.0746 vs.\ 0.0930), the latter matching the baseline RMSRE of the initial rate coefficient estimates (0.0928). Under high noise (variability by a factor of 2), MCMC demonstrates substantially better robustness: on concentration data, MCMC achieves 0.0768 compared to 0.8840 for the Neural Network, and on mass spectrum data, MCMC achieves 0.2921 versus 1.3058 for the Neural Network. The concentration result remains below the baseline RMSRE of 0.0928, whereas the mass-spectral result exceeds the baseline but remains far below the Neural Network error. These results suggest that while the Neural Network excels under low-noise conditions, MCMC is more reliable when measurement
uncertainty is large.

\begin{table}[!ht]
\centering
\caption{Overall RMSRE comparison of MCMC and Neural Network across noise levels. The MCMC values are computed from posterior median rate estimates. The baseline RMSRE denotes the value between the initial  and true coefficient estimates.}
\label{tab:rmsre}
\setlength{\tabcolsep}{8pt}
\renewcommand\arraystretch{1.3}
\resizebox{\columnwidth}{!}{
\begin{tabular}{llcccccc}
\toprule
\multirow{2}{*}{\textbf{Measured data}} 
  & \multirow{2}{*}{\textbf{Baseline}} 
  & \multicolumn{2}{c}{\textbf{noise-free case}} 
  & \multicolumn{2}{c}{\textbf{low-noise case}} 
  & \multicolumn{2}{c}{\textbf{high-noise case}} \\
\cmidrule(lr){3-4} \cmidrule(lr){5-6} \cmidrule(lr){7-8}
  & & \textbf{MCMC} & \textbf{NN} 
    & \textbf{MCMC} & \textbf{NN} 
    & \textbf{MCMC} & \textbf{NN} \\
\midrule
Concentration 
  & \cellcolor{baselineyellow}0.0928& 0.0120 
  & \cellcolor{bestgreen}0.000147& 0.0146 
  & \cellcolor{bestgreen}0.0114& \cellcolor{bestgreen} 0.0768\phantom{0} 
  & 0.884\\
Mass spectrum  
  & \cellcolor{baselineyellow}0.0928& 0.0780 
  & \cellcolor{bestgreen}0.0402& \cellcolor{baselineyellow} 0.0930\phantom{0}
  & \cellcolor{bestgreen}0.0746& \cellcolor{bestgreen}0.2921 
  & 1.306\\
\bottomrule
\end{tabular}}
\end{table}

\begin{figure}
    \centering
    \includegraphics[width=1\linewidth]{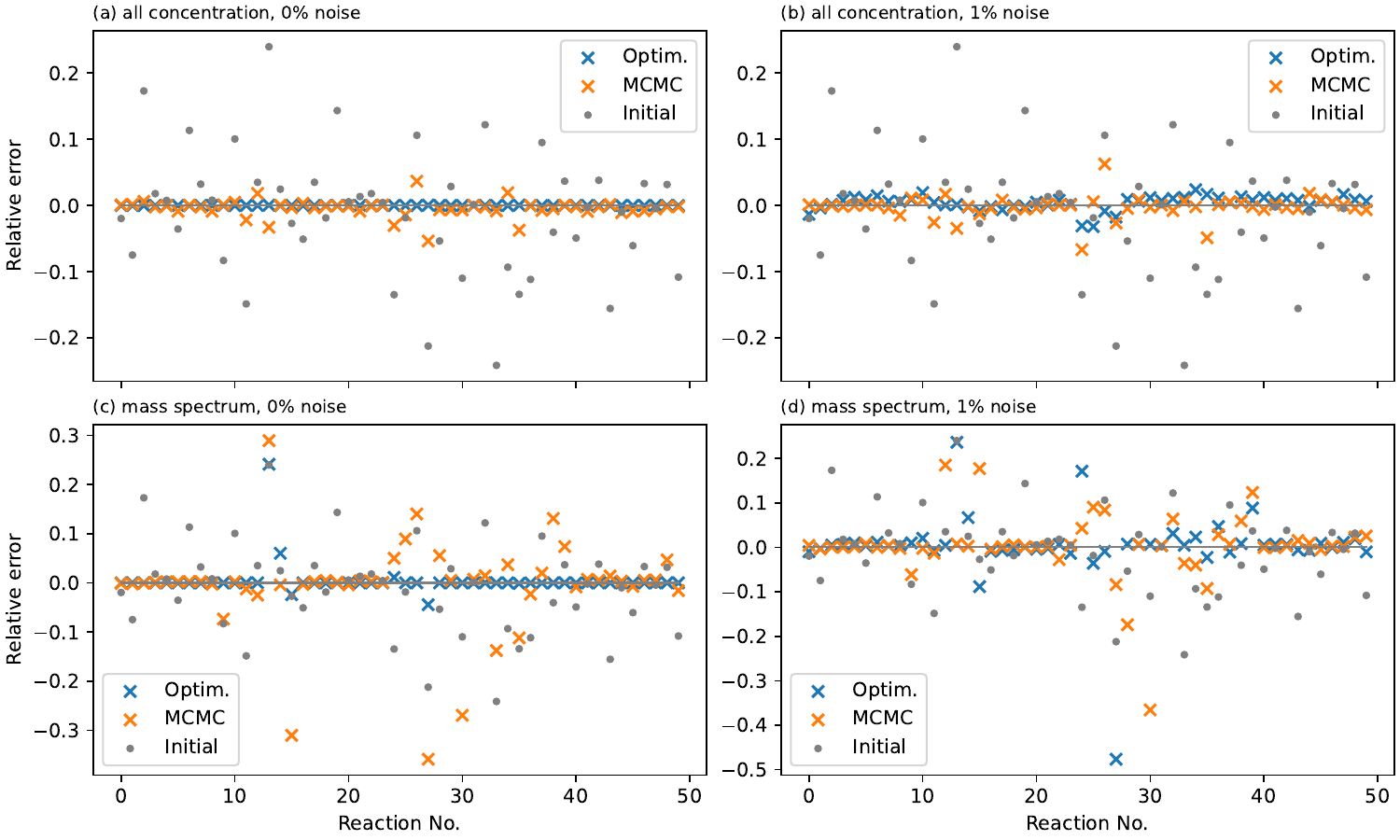}
    \caption{Log relative error of estimated rate coefficients}
    \label{fig:est_k_all}
\end{figure}

In the following sections, we will separately discuss ablation studies of the proposed Neural Networks and MCMC methods.

\subsection{ODE-constrained optimisation}

\input{Section/03_1_result_ODE}

\subsection{DRAM-MCMC posterior inference}

Although six observation cases were considered in the overall comparison, the detailed DRAM-MCMC posterior analysis focuses on three representative cases. These cases were selected to span the main information regimes relevant to the inverse problem: an idealized direct-concentration case, a low-noise mass-spectral case, and a high-noise mass-spectral case. The direct-concentration case provides an upper-limit benchmark for parameter recoverability, whereas mass-spectral aggregation introduces a progressively more information-limited observation scenario. The high-noise mass-spectral case is emphasized as a deliberately pessimistic stress test that may be relevant to weak mass channels near the instrumental background, rather than as a general error model for high-quality mass-spectrometric measurements.

The first case used direct concentration observations at five recorded time points. The second case used synthetic mass spectra with low-noise, where species with the same nominal mass were aggregated before comparison with the observations. Finally, the third case used the same mass-spectral observation operator, but with high-noise. In all cases the synthetic data were generated with true rate multipliers equal to one, so posterior concentration around unity provides a direct measure of rate-coefficient recoverability.

Posterior summaries were computed after discarding the first 20\,\% of each chain as burn-in, leaving $8.0 \times 10^5$ samples per case. Representative retained-chain diagnostics for both observation representations at all three noise levels are provided in Supplementary Figures~S2--S4, with the chain configuration summarized in Supplementary Table~S4. The concentration case gave the tightest posterior constraints (Table~\ref{tab:mcmc_case_summary}). The median width of the parameter-wise 95\,\% credible intervals was 0.015 decades in $\log_{10}$ multiplier space, and the posterior medians ranged only from 0.937 to 1.069. The true multiplier value was contained in the 95\,\% credible interval for 45 of the 50 inferred parameters. Thus, even with only five recorded time points, direct species-level concentration data retained enough information to constrain most of the selected reaction rates closely.

\begin{table}[ht]
\centering
\small
\setlength{\tabcolsep}{5pt}
\caption{Posterior constraints from the three DRAM-MCMC cases after discarding the first 20\,\% of each chain as burn-in. The CI width is reported as the median parameter-wise 95\,\% credible-interval width in $\log_{10}$ multiplier space. The posterior median range is reported in rate-multiplier space. Coverage gives the number of inferred parameters whose 95\,\% credible interval contains the true multiplier value of one.}
\label{tab:mcmc_case_summary}
\begin{tabular}{@{}lccc@{}}
\toprule
Case & 95\% CI width & Median range & Coverage \\
\midrule
Concentration & 0.015 & 0.937--1.069 & 45/50 \\
Mass spectra, low noise & 0.047 & 0.703--1.246 & 47/50 \\
Mass spectra, high noise & 0.613 & 0.294--2.472 & 42/50 \\
\bottomrule
\end{tabular}
\end{table}

The effect of the observation operator is visible when the concentration case is compared with the low-noise mass-spectral case. Although both cases used the same nominal noise level, replacing individual concentrations with mass spectra increased the median credible-interval width from 0.015 to 0.047 decades. This broadening reflects the information loss caused by mass aggregation: multiple species can contribute to the same mass channel, allowing compensating changes in different reaction rates to produce similar spectra. Despite this degeneracy, the low-noise mass-spectral case remained strongly informative. Its posterior medians stayed close to the true value for most parameters, and the 95\,\% credible intervals contained the true multiplier for 47 of the 50 parameters.

The high-noise mass-spectral case showed a qualitatively larger loss of identifiability. At a log-space noise standard deviation of $\sigma=\log(2)$, the median parameter-wise 95\,\% credible-interval width increased to 0.613 decades, and the posterior medians spanned the substantially broader range 0.294--2.472. Only 11 of the 50 parameters had 95\,\% credible intervals contained within a factor of two of the true value, compared with all 50 parameters in the concentration case and 49 parameters in the low-noise mass-spectral case (this demonstrates that the combination of mass aggregation and large measurement uncertainty permits broad compensating changes among reaction rates while preserving agreement with the observations). Complete retained traces for all 50 rate multipliers in the high-noise mass-spectral case are provided in Supplementary Figure~S5, and the corresponding parameter-wise posterior medians and 95\,\% credible intervals are reported in Supplementary Table~S5.

\begin{figure}[ht!]
\centering
\includegraphics[width=0.86\textwidth]{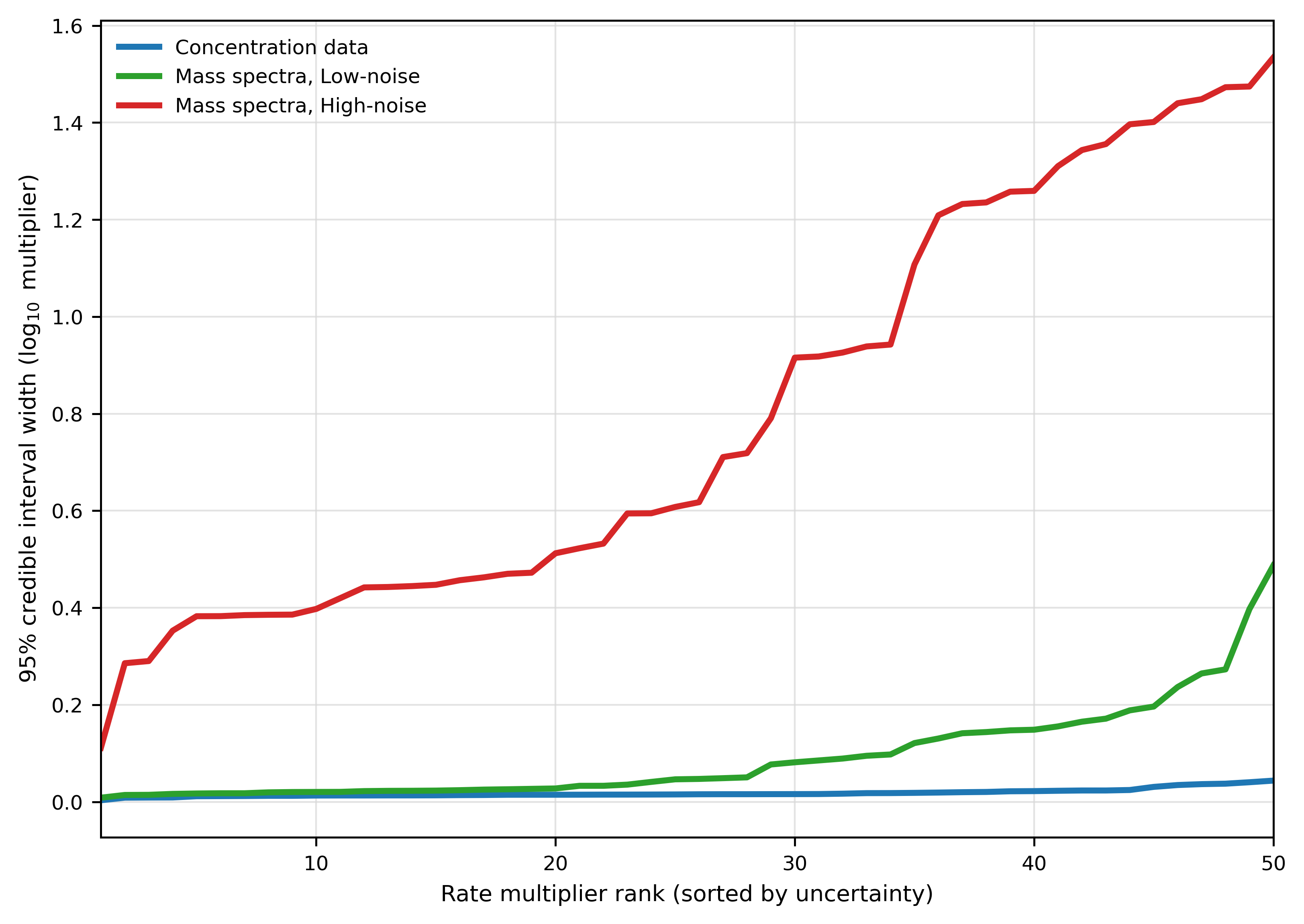}
\caption{Sorted widths of the parameter-wise 95\,\% credible intervals for the three MCMC cases. Direct concentration observations yield the narrowest posterior intervals. Mass-spectral aggregation broadens the posterior under low-noise conditions, while the high-noise mass-spectral case produces substantially weaker constraints for many parameters.}
\label{fig:mcmc_uncertainty_widths}
\end{figure}

Figure~\ref{fig:mcmc_uncertainty_widths} summarizes the full distribution of posterior uncertainty across parameters. The concentration and low-noise mass-spectral cases show relatively compact uncertainty profiles, whereas the high-noise mass-spectral case is shifted upward by more than an order of magnitude for many parameters. The weakest constraints in the high-noise case approach 1.5 decades, corresponding to uncertainty of more than one order of magnitude in the rate multiplier. However, not all parameters become unconstrained: several retain substantially narrower intervals, indicating that some reaction rates remain identifiable because they control features of the observations that cannot be reproduced by compensating changes elsewhere in the mechanism.

\begin{figure}[ht!]
\centering
\includegraphics[width=0.95\textwidth]{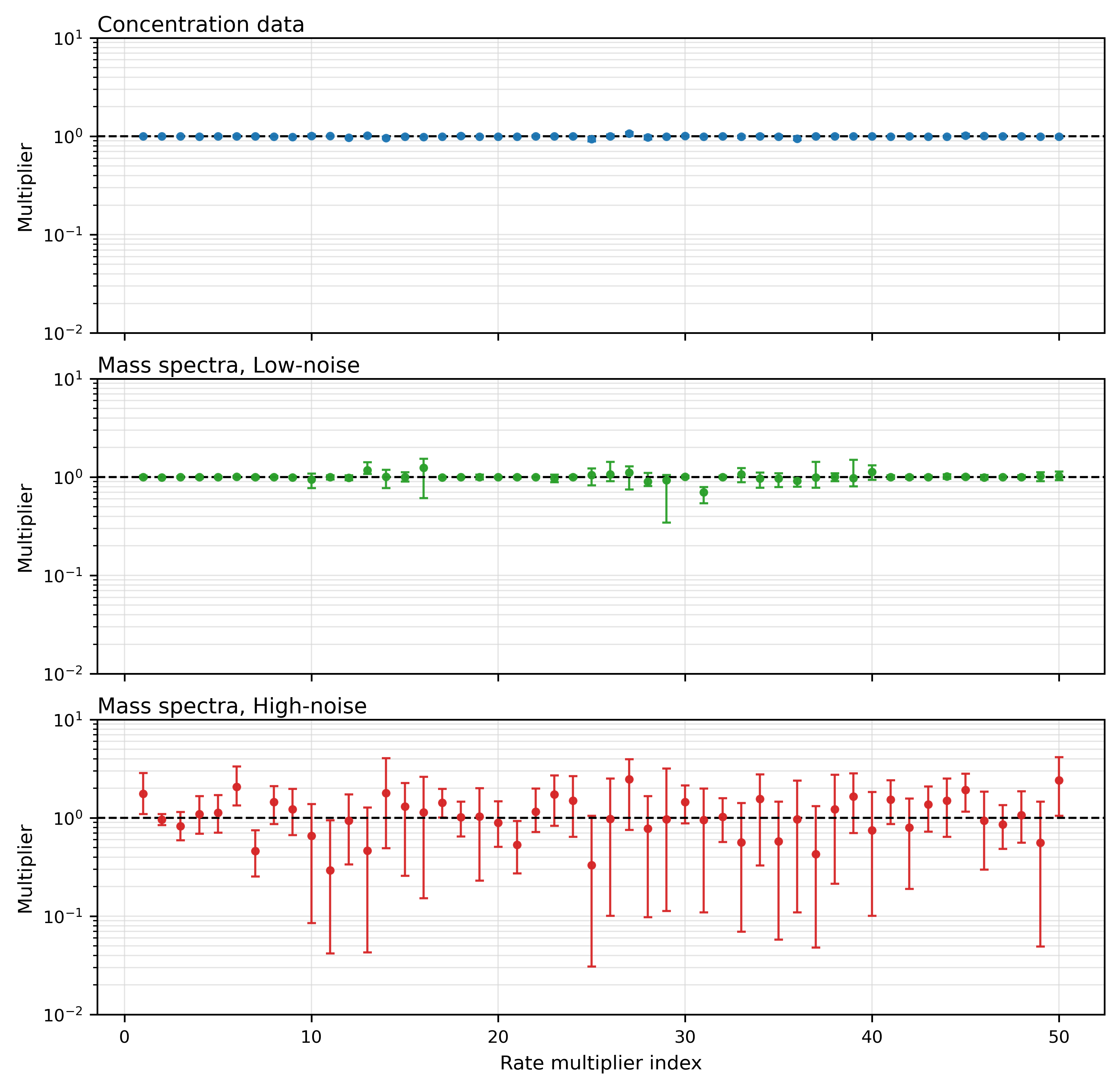}
\caption{Posterior medians and 95\,\% credible intervals for all 50 rate multipliers. The dashed line denotes the true multiplier value of one. The concentration and low-noise mass-spectral cases constrain most parameters tightly around the true value, whereas the high-noise mass-spectral case produces broad intervals and stronger parameter-to-parameter variation in posterior uncertainty.}
\label{fig:mcmc_parameter_intervals}
\end{figure}

The parameter-wise credible intervals in Fig.~\ref{fig:mcmc_parameter_intervals} show that the loss of identifiability is not uniform across the mechanism. In the concentration case, nearly all posterior intervals are narrow and centered close to the true multiplier. The low-noise mass-spectral case preserves this structure for many parameters but introduces noticeably wider intervals for a subset of rates. In the high-noise mass-spectral case, several intervals span a large fraction of the prior-supported range, while others remain moderately localized. This pattern is consistent with a sloppy inverse problem in which only certain combinations of reaction rates are strongly constrained by the available observations.

\begin{figure}[ht!]
\centering
\includegraphics[width=0.82\textwidth]{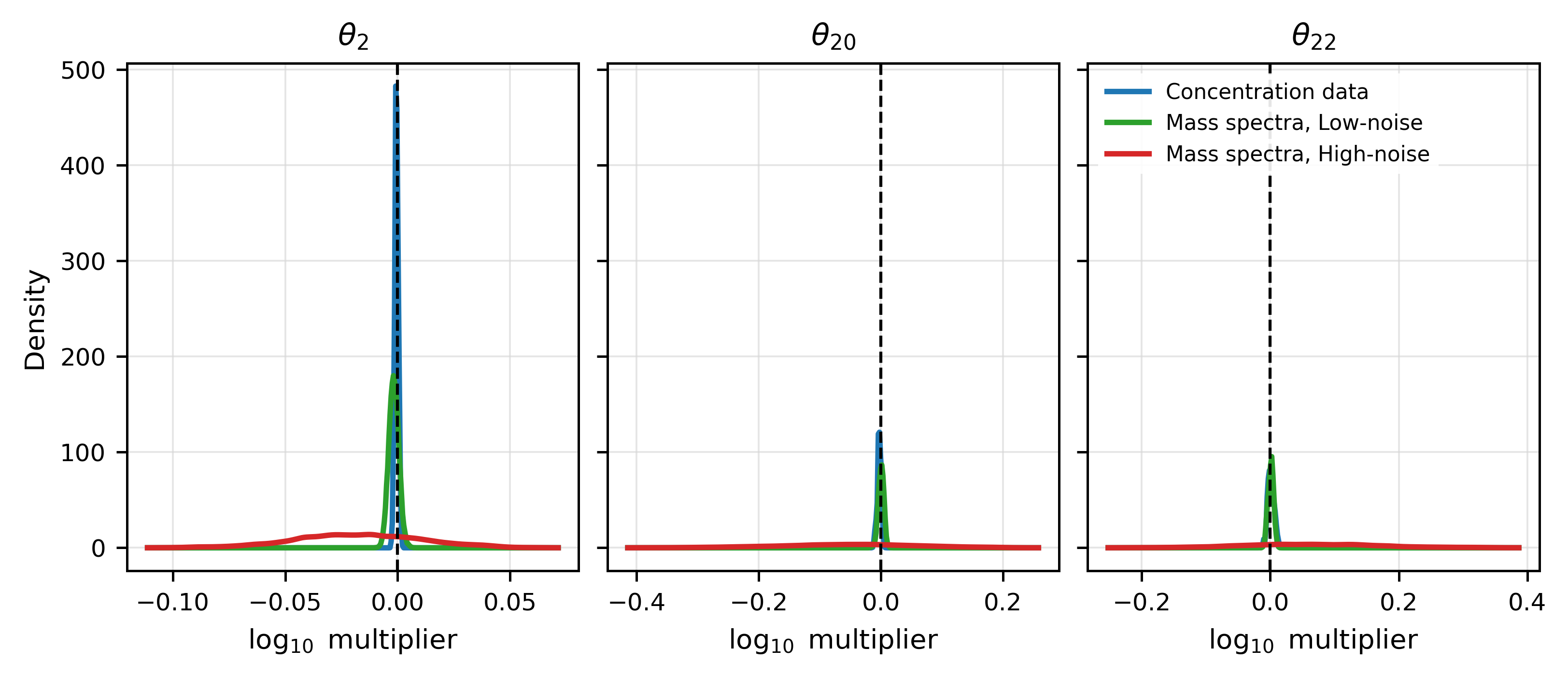}
\caption{Representative marginal posterior distributions for selected rate multipliers in $\log_{10}$ space. The vertical dashed line marks the true value. The examples illustrate the transition from sharply localized posteriors in the concentration and low-noise mass-spectral cases to broader and more weakly identified posteriors in the high-noise mass-spectral case.}
\label{fig:mcmc_representative_posteriors}
\end{figure}

Representative marginal posterior distributions are shown in Fig.~\ref{fig:mcmc_representative_posteriors}. The dashed vertical line at zero corresponds to the true multiplier value ($\log_{10}(1)=0$). For the concentration and low-noise mass-spectral cases, the selected posteriors are sharply localized near the true value, indicating that these parameters are well constrained by the observations. In contrast, the high-noise mass-spectral case produces much broader marginal posteriors, and for some parameters the posterior mass shifts away from the true value. This indicates weaker identifiability under high observational noise and mass aggregation. The figure therefore illustrates why a single best-fitting parameter vector is insufficient: several rate-multiplier combinations can remain plausible, especially when the observations are noisy and mass-aggregated.

\subsubsection{Posterior Predictive Validation of Mass Spectra}

The posterior uncertainty described above raises an important validation question: whether the range of plausible rate multipliers still reproduces the observable mass spectrum. To test this, we propagated 200 posterior rate-multiplier samples through the forward model and aggregated the resulting species concentrations into nominal mass channels. For each mass channel, the posterior predictive distribution was summarized by its mean and 90\,\% interval.
This posterior predictive check evaluates the ensemble of parameter sets sampled by MCMC, rather than only a single fitted parameter vector. It therefore tests whether the inferred posterior distribution is consistent with the synthetic observation while preserving the uncertainty caused by mass aggregation and measurement noise.

\begin{figure}[ht]
\centering
\includegraphics[width=\textwidth]{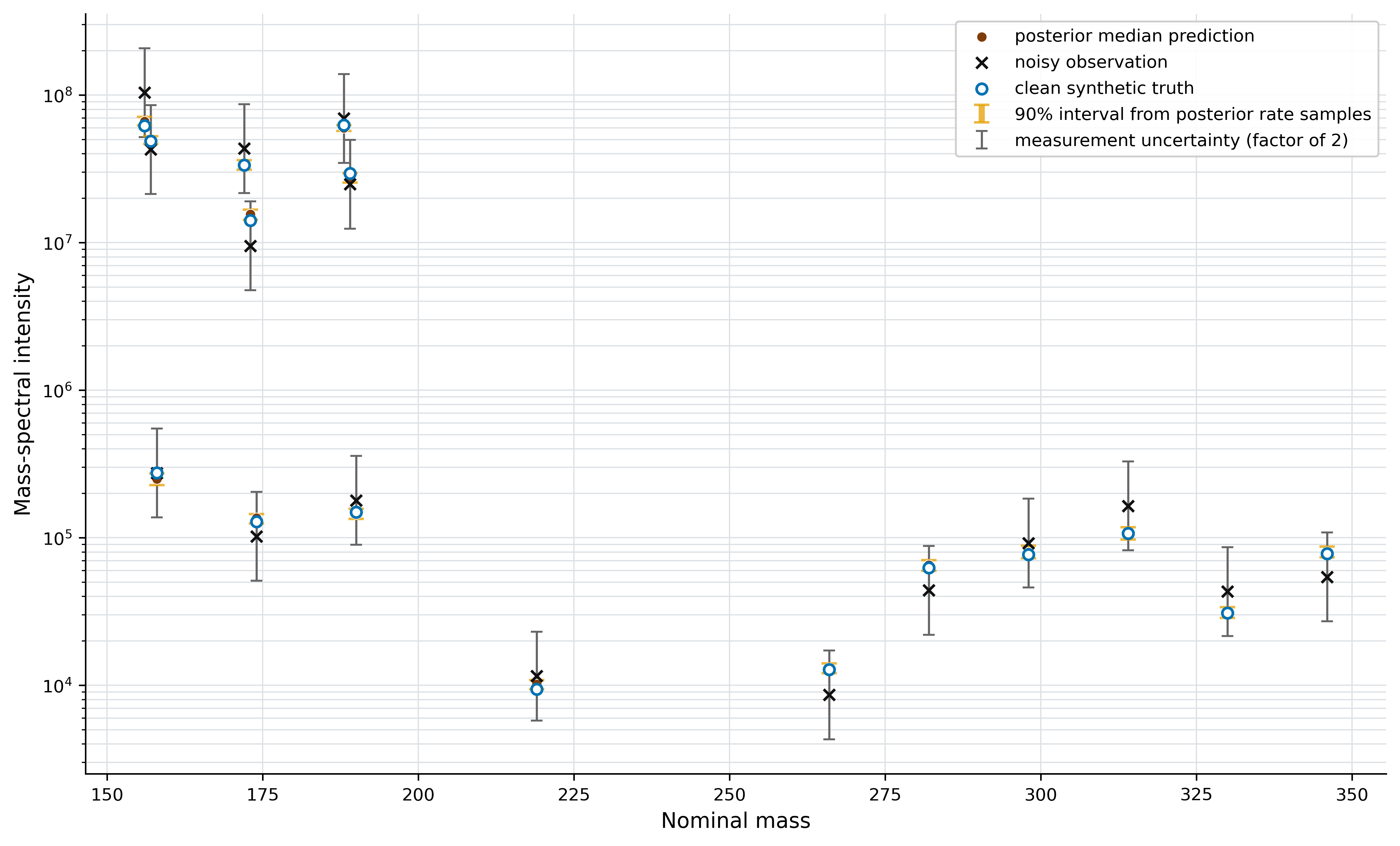}
\caption{
Posterior predictive check for the high-noise mass-spectral case. The black crosses show the noisy observation, the blue open circles show the clean synthetic spectrum, and the brown markers show the posterior mean prediction from 200 posterior samples. Error bars denote the 90\,\% posterior predictive interval for each nominal mass channel.
}
\label{fig:spectra_validation}
\end{figure}

Figure~\ref{fig:spectra_validation} shows that the posterior predictive mean closely follows the clean synthetic spectrum across the observed mass range, while the noisy observations scatter around it because of the imposed measurement noise. Across the 200 posterior predictive simulations, the mean MSLE was 0.00815 relative to the clean synthetic spectrum and 0.130 relative to the noisy observation. The 90\,\% predictive intervals are narrow for most mass channels, indicating that the observable mass spectrum is stable across the sampled posterior parameter sets. This behaviour is consistent with the broad parameter intervals shown above: individual reaction rates can remain weakly identified, but their compensating effects still produce a well-constrained mass-spectral prediction.

%% file: Section/03_1_result_ODE.tex
We apply the ODE-constrained optimisation step of the SPIN-ODE framework to recover the uncertain rate coefficients by minimising the trajectory misfit. The method is evaluated under varying prior uncertainty in the rate coefficients and under different observation conditions, namely full concentrations versus mass spectra and increasing noise levels. During optimisation, we record the MSLE of both the trajectory fitting objective and the estimated rate coefficients.

\begin{figure}[htbp]
    \centering
    \begin{subfigure}[t]{0.48\textwidth}
        \centering
        \includegraphics[width=\textwidth]{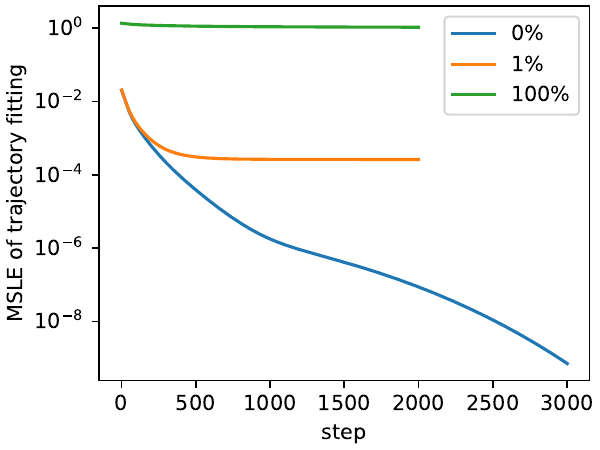}
        \caption{Trajectory fitting against all concentrations.}
        \label{fig:loss-conc}
    \end{subfigure}
    \hfill
    \begin{subfigure}[t]{0.48\textwidth}
        \centering
        \includegraphics[width=\textwidth]{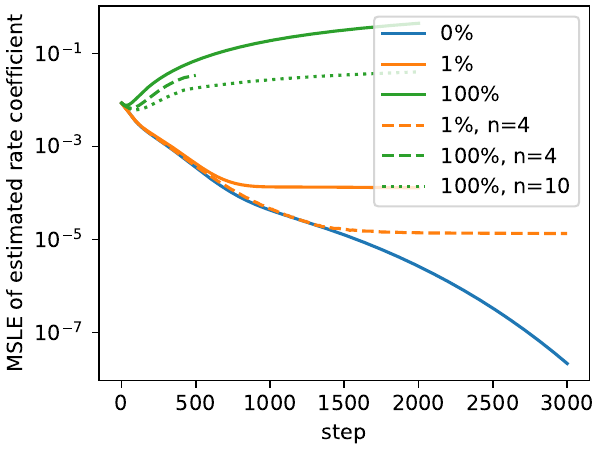}
        \caption{Rate coefficient optimisation against all concentrations.}
        \label{fig:k_err-conc}
    \end{subfigure}

    \vspace{1ex}   

    \begin{subfigure}[t]{0.48\textwidth}
        \centering
        \includegraphics[width=\textwidth]{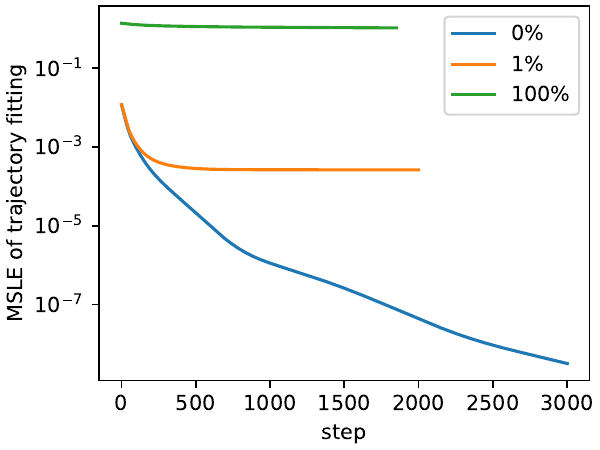}
        \caption{Trajectory fitting against mass spectrum.}
        \label{fig:loss-mass}
    \end{subfigure}
    \hfill
    \begin{subfigure}[t]{0.48\textwidth}
        \centering
        \includegraphics[width=\textwidth]{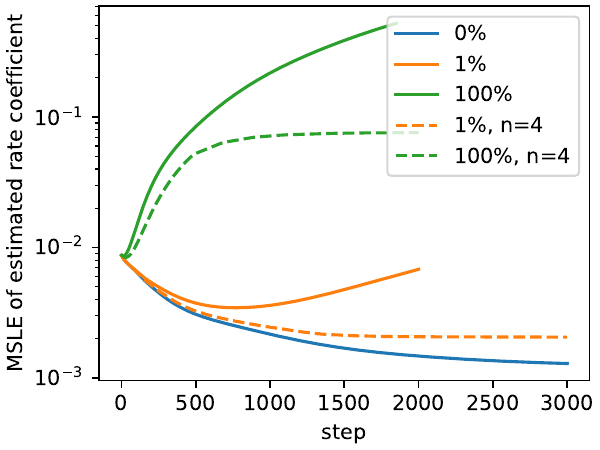}
        \caption{Rate coefficient optimisation against mass spectrum.}
        \label{fig:k_err-mass}
    \end{subfigure}

    \caption{Optimisation convergence. Panels~\ref{fig:loss-conc} and~\ref{fig:loss-mass} show the trajectory-fitting objective, and panels~\ref{fig:k_err-conc} and~\ref{fig:k_err-mass} the corresponding error of the estimated rate coefficients. Solid lines use a single noisy realisation; dashed lines use stochastic optimisation over $n=4$ noisy samples.}
    \label{fig:opt_convergence}
\end{figure}

Figures~\ref{fig:loss-conc} and~\ref{fig:loss-mass} show the convergence of the objective when optimising against the concentrations and the mass spectrum, respectively. In both cases the objective converges for the 0\% and 1\% noise levels. At 0\% noise the objective decreases throughout the 3000 steps, reaching values of order $10^{-9}$, whereas at 1\% noise it plateaus earlier, at $2.6\times10^{-4}$. At high-noise the objective starts roughly one to two orders of magnitude higher and barely decreases, indicating that the residual is dominated by the observation noise rather than by uncertainty in the rate coefficients. Consequently, optimisation of the rate coefficients has much less effect on the objective than under the other noise conditions. Figures~\ref{fig:k_err-conc} and~\ref{fig:k_err-mass} show that the rate coefficients are recovered accurately under noise-free and low-noise conditions when fitting the concentrations, and at noise-free when fitting the mass spectrum. For other cases, the rate-coefficient error decreases first but then turns upward and diverges.

Under noisy observations, where each trajectory is perturbed using an independent realization of the multiplicative noise model defined in the Methods, the recovery degrades. Stochastic gradient descent mitigates this by optimising against multiple independent noise realisations. The standard error of the mean (SEM) decreases with the number of samples $n$ as $\text{SEM}=\sigma/\sqrt{n}$. To exploit this, we generate $n$ noisy copies of the original observations, each with the same mean and variance. After omitting the first of the five recorded time points, the resulting dataset has dimensions $\mathbb{R}^{n\times10\times4\times50}$. Gradient descent is then applied over mini-batches of dimensions $\mathbb{R}^{B\times10\times4\times50}$, where $B\leq n$. The dashed lines in Figure~\ref{fig:opt_convergence} show the improved convergence of this multi-sample optimisation, and Table~\ref{tab:k_extra_noise} compares it with the single-realisation results of Table~\ref{tab:rmsre}. At low noise, the rate-coefficient error decreases by 68\% for the concentrations and 36\% for the mass spectrum, and the divergence in the mass-spectrum case is removed. Under high-noise conditions, the divergence is reduced but not eliminated. In the high-noise concentration case, optimisation with four noise samples terminates early, whereas a separate test with ten samples reduces the error from 0.884 to 0.217. The improvement does not scale indefinitely with the number of samples: adding further samples does not necessarily improve on the results shown, although the outcome remains better than the baseline. A complete comparison across sample counts, together with the corresponding trajectory-fitting objectives, is provided in Supplementary Figure~S1.

\begin{table}[!ht]
\centering
\caption{RMSRE of estimated rate coefficients with additional noise samples.}
\label{tab:k_extra_noise}
\begin{tabular}{cccccc}
\hline
\multirow{2}{*}{} & \multirow{2}{*}{Baseline} & \multicolumn{2}{c}{Low-noise} & \multicolumn{2}{c}{High-noise} \\ \cline{3-6} 
                  &                           & n=1        & n=4        & n=1          & n=4        \\ \hline
Conc.             & 0.0928& 0.0114     & 0.00367    & 0.884        & N/A        \\
Mass              & 0.0928& 0.0746     & 0.0478     & 1.306& 0.289      \\ \hline
\end{tabular}
\end{table}

\subsubsection{Ablation study of ODE-constrained optimisation}
The ODE-constrained optimisation integrates the time derivatives of the kinetic model, so that the rate coefficients are constrained simultaneously by the kinetic rate law and by the integrated ODE dynamics. To isolate the role of the ODE constraint, we remove the integration and instead optimise the kinetic model directly against the time derivatives obtained from the observed trajectories using a finite-difference approximation. This is equivalent to the second step of the SPIN-ODE framework without the dense neural interpolation.

\begin{figure}
    \centering
    \includegraphics[width=1\linewidth]{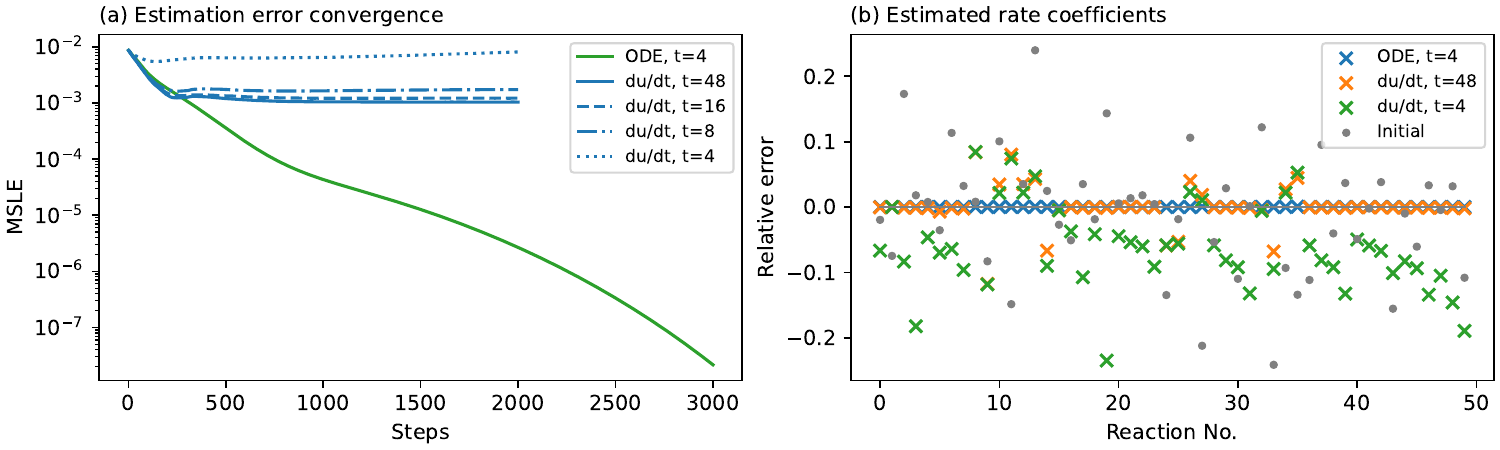}
    \caption{Comparison of optimisation against ODE trajectory and finite difference derivative, using 0\% noise concentration observations. $t$ is the number of time steps sampled from the trajectories.}
    \label{fig:opt_dydt}
\end{figure}
The derivative optimisation does not benefit from additional trajectories with different initial concentrations and performs better with one trajectory than with ten. Therefore, Figure~\ref{fig:opt_dydt} shows ODE optimisation against ten trajectories and derivative optimisation against one trajectory. The derivative optimisation is highly sensitive to temporal sampling density. Using the same four time points as the ODE optimisation, it barely improves the rate coefficients, most of which are underestimated, reflecting the severe information loss caused by finite differencing of sparse trajectories. Increasing the sampling density to 48 time points correctly recovers 35 of the 50 rate coefficients. The reactions that still fail, including reactions 8--13, are intermediate propagation steps in the radical system. This suggests that finite differences cannot adequately capture the comparatively fast dynamics of the stiff system. By contrast, ODE-constrained optimisation correctly recovers all 50 rate coefficients.

%% file: Section/04_Conclusion.tex
\section{Conclusion}
This study evaluated reaction-rate optimisation for a toy-case autoxidation mechanism using synthetic observations with known ground truth. The comparison between ODE-constrained neural-network optimisation and DRAM-MCMC posterior inference showed that the relative performance of the two approaches depends strongly on the information content and uncertainty of the observations. When the synthetic observations were noise-free or only weakly perturbed, the neural-network approach recovered the rate coefficients with low RMSRE and outperformed the posterior-mean MCMC estimates. Under high observational noise, defined by a log-space standard deviation of $\log(2)$ and approximately S/N${}=1$, DRAM-MCMC was substantially more robust, particularly for the synthetic mass-spectral observations.

The posterior analysis further showed that rate-coefficient recovery is influenced not only by the noise level but also by the observation representation. Direct concentration observations provided relatively strong constraints on most of the inferred reaction rates. In the synthetic mass-spectral representation, however, the concentrations of species with the same nominal molecular mass were summed into a single nominal-mass channel. This species-to-nominal-mass aggregation removed species-level information and broadened the posterior distributions even under low noise. In the high-noise mass-spectral case, many individual rate coefficients became weakly identifiable, resulting in broad credible intervals and compensating parameter combinations. Nevertheless, posterior predictive simulations reproduced the principal features of the observable mass spectrum, indicating that the nominal-mass distribution can be better constrained than the individual reaction rates that produce it.

The high-noise case should be interpreted as a deliberately severe stress test rather than as a uniform representation of high-quality mass-spectrometric measurements. Contemporary chemical-ionisation mass-spectrometry studies commonly define detection limits using a signal-to-noise ratio of approximately three, indicating that signals with S/N${}\approx1$ would generally be considered close to or below the reliable detection threshold \citep{Yuan2016ToFCIMS,Song2024ICIMS}. The high-noise case considered here is therefore pessimistic for strong signals that are well above the detection limit, although it may be representative of weak mass channels close to the instrumental background. In addition, reported uncertainties in the quantification of highly oxygenated organic molecules can range from approximately $\pm50\,\%$ to a factor of two because of calibration uncertainty and compound-dependent ionisation sensitivity \citep{Rissanen2019MION}. Such calibration uncertainties are systematic and differ from the independent multiplicative perturbations imposed in the present synthetic experiment. Consequently, the high-noise case should be regarded as a robustness test representing a difficult measurement regime rather than as a detailed error model for a particular instrument.

These results highlight the complementary roles of the two inference approaches. ODE-constrained neural-network optimisation is effective when the observations are sufficiently informative and a point estimate is the primary objective. DRAM-MCMC is computationally more demanding, but it provides uncertainty quantification and exposes parameter non-uniqueness, which are particularly important when the inverse problem involves noisy, sparse, or nominal-mass-aggregated observations. For atmospheric autoxidation mechanisms, this distinction is important because observable chemical signatures can remain well reproduced even when some of the internal reaction rate coefficients are not uniquely identifiable.

Overall, the toy-case results suggest that rate-coefficient estimation for newly generated chemical mechanisms can benefit from combining efficient point-estimation methods with Bayesian uncertainty analysis. Such a workflow can identify plausible rate coefficients, determine which parameters are genuinely constrained by the available observations, and avoid overinterpreting a single fitted mechanism when multiple rate combinations remain consistent with the data. Future work should evaluate these methods using instrument-specific error models that separately account for counting noise, background variability, calibration uncertainty, compound-dependent sensitivity, and detection limits before applying the framework to experimental mass-spectrometric observations.

%% file: Section/0_1_Environmental_Significance.tex
\section*{Environmental Significance Statement}
Uncertainty in reaction-rate coefficients limits the reliability of atmospheric chemical mechanisms and their predictions of chemical composition. Estimating these coefficients is challenging because atmospheric observations may be noisy, incomplete, or aggregated by measurement instruments. This study evaluates optimisation and Bayesian inference methods for recovering reaction-rate coefficients from concentration and mass-spectral observations using a controlled toy mechanism. The results demonstrate how measurement type and noise affect the findings and their implications.

%% file: Section/0_2_Data_Availability.tex
\section*{Data availability}
The data and code supporting this study are available from the corresponding author upon reasonable request.

%% file: Section/0_3_Author_Contribution.tex
\section*{Author contributions}

VA, HH, LP and MB designed the research. VA, WP, ZL, PC and HH developed the computational workflow, conducted the numerical experiments and analysed the results. VA and HH developed and implemented the parameter-estimation methods. WP and ZL developed and implemented the SPIN-ODE method. AR developed and implemented the dimensionality-reduction method. VA, WP, HH, TA, ZF and MB prepared the original manuscript. All authors reviewed and edited the manuscript and approved the final version.

%% file: Section/acknowledgements.tex
\section*{Acknowledgements}
We thank the City of Lahti for financial support and CSC -- IT Center for Science, Finland, for providing computational resources. AR acknowledges support from the Deutsche Forschungsgemeinschaft (DFG, German Research Foundation; grant 577175348). WP acknowledges financial support from the Finnish Cultural Foundation, Päijät-Häme Regional Fund (grant 70251819). We thank Professor Matti Rissanen of Tampere University for constructive discussions, particularly regarding the design of the toy-case mechanism and the assignment of its reference reaction-rate coefficients.


%% file: main.bib
@article{Atkinson2003VOCDegradation,
    author  = {Atkinson, Roger and Arey, Janet},
    title   = {Atmospheric Degradation of Volatile Organic Compounds},
    journal = {Chemical Reviews},
    year    = {2003},
    volume  = {103},
    number  = {12},
    pages   = {4605--4638},
    doi     = {10.1021/cr0206420}
  }

@book{Seinfeld2016AtmosphericChemistry,
    author    = {Seinfeld, John H. and Pandis, Spyros N.},
    title     = {Atmospheric Chemistry and Physics: From Air Pollution to
    Climate Change},
    edition   = {3},
    publisher = {John Wiley \& Sons},
    address   = {Hoboken, New Jersey},
    year      = {2016},
    isbn      = {9781118947401}
  }

@article{Bianchi2019HOM,
    author  = {Bianchi, Federico and Kurt{\'e}n, Theo and Riva, Matthieu and
    Mohr, Claudia and Rissanen, Matti P. and Roldin, Pontus and Berndt, Torsten
    and Crounse, John D. and Wennberg, Paul O. and Mentel, Thomas F. and Wildt,
    J{\"u}rgen and Junninen, Heikki and Jokinen, Tuija and Kulmala, Markku and
    Worsnop, Douglas R. and Thornton, Joel A. and Donahue, Neil and Kjaergaard,
    Henrik G. and Ehn, Mikael},
    title   = {Highly Oxygenated Organic Molecules ({HOM}) from Gas-Phase
    Autoxidation Involving Peroxy Radicals: A Key Contributor to Atmospheric
    Aerosol},
    journal = {Chemical Reviews},
    year    = {2019},
    volume  = {119},
    number  = {6},
    pages   = {3472--3509},
    doi     = {10.1021/acs.chemrev.8b00395}
  }

@article{Ehn2014LowVolatilitySOA,
    author  = {Ehn, Mikael and Thornton, Joel A. and Kleist, Einhard and
    Sipil{\"a}, Mikko and Junninen, Heikki and Pullinen, Iida and Springer,
    Monika and Rubach, Florian and Tillmann, Ralf and Lee, Ben and Lopez-
    Hilfiker, Felipe and Andres, Stefanie and Acir, Ismail-Hakki and Rissanen,
    Matti and Jokinen, Tuija and Schobesberger, Siegfried and Kangasluoma, Juha
    and Kontkanen, Jenni and Nieminen, Tuomo and Kurt{\'e}n, Theo and Nielsen,
    Lasse B. and J{\o}rgensen, Solvejg and Kjaergaard, Henrik G. and
    Canagaratna, Manjula and Dal Maso, Miikka and Berndt, Torsten and Pet{\"a}
    j{\"a}, Tuukka and Wahner, Andreas and Kerminen, Veli-Matti and Kulmala,
    Markku and Worsnop, Douglas R. and Wildt, J{\"u}rgen and Mentel, Thomas F.},
    title   = {A Large Source of Low-Volatility Secondary Organic Aerosol},
    journal = {Nature},
    year    = {2014},
    volume  = {506},
    number  = {7489},
    pages   = {476--479},
    doi     = {10.1038/nature13032}
  }

@article{Guenther2012MEGAN21,
    author  = {Guenther, A. B. and Jiang, X. and Heald, C. L. and
    Sakulyanontvittaya, T. and Duhl, T. and Emmons, L. K. and Wang, X.},
    title   = {The Model of Emissions of Gases and Aerosols from Nature Version
    2.1 ({MEGAN2.1}): An Extended and Updated Framework for Modeling Biogenic
    Emissions},
    journal = {Geoscientific Model Development},
    year    = {2012},
    volume  = {5},
    number  = {6},
    pages   = {1471--1492},
    doi     = {10.5194/gmd-5-1471-2012}
  }

@article{Haario2006DRAM,
    author  = {Haario, Heikki and Laine, Marko and Mira, Antonietta and Saksman,
    Eero},
    title   = {{DRAM}: Efficient Adaptive {MCMC}},
    journal = {Statistics and Computing},
    year    = {2006},
    volume  = {16},
    number  = {4},
    pages   = {339--354},
    doi     = {10.1007/s11222-006-9438-0}
  }

@article{Hallquist2009SOA,
    author  = {Hallquist, M. and Wenger, J. C. and Baltensperger, U. and Rudich,
    Y. and Simpson, D. and Claeys, M. and Dommen, J. and Donahue, N. M. and
    George, C. and Goldstein, A. H. and Hamilton, J. F. and Herrmann, H. and
    Hoffmann, T. and Iinuma, Y. and Jang, M. and Jenkin, M. E. and Jimenez, J.
    L. and Kiendler-Scharr, A. and Maenhaut, W. and McFiggans, G. and Mentel, T.
    F. and Monod, A. and Pr{\'e}v{\^o}t, A. S. H. and Seinfeld, J. H. and
    Surratt, J. D. and Szmigielski, R. and Wildt, J.},
    title   = {The Formation, Properties and Impact of Secondary Organic
    Aerosol: Current and Emerging Issues},
    journal = {Atmospheric Chemistry and Physics},
    year    = {2009},
    volume  = {9},
    number  = {14},
    pages   = {5155--5236},
    doi     = {10.5194/acp-9-5155-2009}
  }

@article{Jimenez2009OrganicAerosols,
    author  = {Jimenez, J. L. and Canagaratna, M. R. and Donahue, N. M. and
    Pr{\'e}v{\^o}t, A. S. H. and Zhang, Q. and Kroll, J. H. and DeCarlo, P. F.
    and Allan, J. D. and Coe, H. and Ng, N. L. and Aiken, A. C. and Docherty, K.
    S. and Ulbrich, I. M. and Grieshop, A. P. and Robinson, A. L. and Duplissy,
    J. and Smith, J. D. and Wilson, K. R. and Lanz, V. A. and Hueglin, C. and
    Sun, Y. L. and Tian, J. and Laaksonen, A. and Raatikainen, T. and
    Rautiainen, J. and Vaattovaara, P. and Ehn, M. and Kulmala, M. and
    Tomlinson, J. M. and Collins, D. R. and Cubison, M. J. and Dunlea, E. J. and
    Huffman, J. A. and Onasch, T. B. and Alfarra, M. R. and Williams, P. I. and
    Bower, K. and Kondo, Y. and Schneider, J. and Drewnick, F. and Borrmann, S.
    and Weimer, S. and Demerjian, K. and Salcedo, D. and Cottrell, L. and
    Griffin, R. and Takami, A. and Miyoshi, T. and Hatakeyama, S. and Shimono,
    A. and Sun, J. Y. and Zhang, Y. M. and Dzepina, K. and Kimmel, J. R. and
    Sueper, D. and Jayne, J. T. and Herndon, S. C. and Trimborn, A. M. and
    Williams, L. R. and Wood, E. C. and Middlebrook, A. M. and Kolb, C. E. and
    Baltensperger, U. and Worsnop, D. R.},
    title   = {Evolution of Organic Aerosols in the Atmosphere},
    journal = {Science},
    year    = {2009},
    volume  = {326},
    number  = {5959},
    pages   = {1525--1529},
    doi     = {10.1126/science.1180353}
  }

@article{McFiggans2019SOAMixtures,
    author  = {McFiggans, Gordon and Mentel, Thomas F. and Wildt, J{\"u}rgen and
    Pullinen, Iida and Kang, Sungah and Kleist, Einhard and Schmitt, Sebastian
    and Springer, Monika and Tillmann, Ralf and Wu, Cheng and Zhao, Defeng and
    Hallquist, Mattias and Faxon, Cynthia and Le Breton, Michael and Hallquist,
    {\AA}sa M. and Simpson, David and Bergstr{\"o}m, Robert and Jenkin, Michael
    E. and Ehn, Mikael and Thornton, Joel A. and Alfarra, M. Rami and Bannan,
    Thomas J. and Percival, Carl J. and Priestley, Michael and Topping, David
    and Kiendler-Scharr, Astrid},
    title   = {Secondary Organic Aerosol Reduced by Mixture of Atmospheric
    Vapours},
    journal = {Nature},
    year    = {2019},
    volume  = {565},
    number  = {7741},
    pages   = {587--593},
    doi     = {10.1038/s41586-018-0871-y}
  }

@article{Roldin2019HOMBoreal,
    author  = {Roldin, Pontus and Ehn, Mikael and Kurt{\'e}n, Theo and Olenius,
    Tinja and Rissanen, Matti P. and Sarnela, Nina and Elm, Jonas and Rantala,
    Pekka and Hao, Liqing and Hyttinen, Noora and Heikkinen, Liine and Worsnop,
    Douglas R. and Pichelstorfer, Lukas and Xavier, Carlton and Clusius, Petri
    and {\"O}str{\"o}m, Emilie and Pet{\"a}j{\"a}, Tuukka and Kulmala,
    Markku and Vehkam{\"a}ki, Hanna and Virtanen, Annele and Riipinen, Ilona and
    Boy, Michael},
    title   = {The Role of Highly Oxygenated Organic Molecules in the Boreal
    Aerosol-Cloud-Climate System},
    journal = {Nature Communications},
    year    = {2019},
    volume  = {10},
    pages   = {4370},
    doi     = {10.1038/s41467-019-12338-8}
  }

@misc{laine2023mcmcstat,
  author       = {Laine, Marko},
  title        = {MCMCSTAT: A MATLAB package for Bayesian analysis using Markov Chain Monte Carlo simulation},
  year         = {2023},
  howpublished = {\url{https://github.com/mjlaine/mcmcstat}},
  note         = {Accessed: 2025-05-22}
}

@Article{Pichelstoerfer,
AUTHOR = {Pichelstorfer, L. and Roldin, P. and Rissanen, M. and Hyttinen, N. and Garmash, O. and Xavier, C. and Zhou, P. and Clusius, P. and Foreback, B. and Golin Almeida, T. and Deng, C. and Baykara, M. and Kurten, T. and Boy, M.},
TITLE = {Towards a mechanistic description of autoxidation chemistry: from precursors to atmospheric implications},
JOURNAL = {EGUsphere},
VOLUME = {2023},
YEAR = {2023},
PAGES = {1--30},
URL = {https://egusphere.copernicus.org/preprints/2023/egusphere-2023-1415/},
DOI = {10.5194/egusphere-2023-1415}
}

@article{smiles,
author = {Weininger, David},
title = {SMILES, a chemical language and information system. 1. Introduction to methodology and encoding rules},
journal = {Journal of Chemical Information and Computer Sciences},
volume = {28},
number = {1},
pages = {31-36},
year = {1988},
doi = {10.1021/ci00057a005},

URL = { 
    
        https://doi.org/10.1021/ci00057a005
    
    

},
eprint = { 
    
        https://doi.org/10.1021/ci00057a005
    
    

}

}

@Article{arcabox,
AUTHOR = {Clusius, P. and Xavier, C. and Pichelstorfer, L. and Zhou, P. and Olenius, T. and Roldin, P. and Boy, M.},
TITLE = {Atmospherically Relevant Chemistry and Aerosol box model -- ARCA box (version 1.2)},
JOURNAL = {Geoscientific Model Development},
VOLUME = {15},
YEAR = {2022},
NUMBER = {18},
PAGES = {7257--7286},
URL = {https://gmd.copernicus.org/articles/15/7257/2022/},
DOI = {10.5194/gmd-15-7257-2022}
}

@Article{adchem,
AUTHOR = {Roldin, P. and Swietlicki, E. and Schurgers, G. and Arneth, A. and Lehtinen, K. E. J. and Boy, M. and Kulmala, M.},
TITLE = {Development and evaluation of the aerosol dynamics and gas phase chemistry model ADCHEM},
JOURNAL = {Atmospheric Chemistry and Physics},
VOLUME = {11},
YEAR = {2011},
NUMBER = {12},
PAGES = {5867--5896},
URL = {https://acp.copernicus.org/articles/11/5867/2011/},
DOI = {10.5194/acp-11-5867-2011}
}

@Article{adcham,
AUTHOR = {Roldin, P. and Eriksson, A. C. and Nordin, E. Z. and Hermansson, E. and Mogensen, D. and Rusanen, A. and Boy, M. and Swietlicki, E. and Svenningsson, B. and Zelenyuk, A. and Pagels, J.},
TITLE = {Modelling non-equilibrium secondary organic aerosol formation and evaporation with the aerosol dynamics, gas- and particle-phase chemistry kinetic multilayer model ADCHAM},
JOURNAL = {Atmospheric Chemistry and Physics},
VOLUME = {14},
YEAR = {2014},
NUMBER = {15},
PAGES = {7953--7993},
URL = {https://acp.copernicus.org/articles/14/7953/2014/},
DOI = {10.5194/acp-14-7953-2014}
}

@misc{Peng2025SPINODE,
    author        = {Peng, Wenqing and Liu, Zhi-Song and Boy, Michael},
    title         = {{SPIN-ODE}: Stiff Physics-Informed Neural {ODE} for
    Chemical Reaction Rate Estimation},
    year          = {2025},
    eprint        = {2505.05625},
    archivePrefix = {arXiv},
    primaryClass  = {cs.LG},
    doi           = {10.48550/arXiv.2505.05625},
    note          = {Accepted at the European Conference on Artificial
    Intelligence (ECAI) 2025}
  }

@Article{Lukas_CONSTRAINTS,
AUTHOR = {Pichelstorfer, L. and O'Meara, S. P. and McFiggans, G. B.},
TITLE = {Theory informed, experiment based, constraint on the rate of autoxidation chemistry -- An analytical approach},
JOURNAL = {Aerosol Research Discussions},
VOLUME = {2024},
YEAR = {2024},
PAGES = {1--20},
URL = {https://ar.copernicus.org/preprints/ar-2024-40/},
DOI = {10.5194/ar-2024-40}
}

@Article{kanakaidou,
AUTHOR = {Kanakidou, M. and Seinfeld, J. H. and Pandis, S. N. and Barnes, I. and Dentener, F. J. and Facchini, M. C. and Van Dingenen, R. and Ervens, B. and Nenes, A. and Nielsen, C. J. and Swietlicki, E. and Putaud, J. P. and Balkanski, Y. and Fuzzi, S. and Horth, J. and Moortgat, G. K. and Winterhalter, R. and Myhre, C. E. L. and Tsigaridis, K. and Vignati, E. and Stephanou, E. G. and Wilson, J.},
TITLE = {Organic aerosol and global climate modelling: a review},
JOURNAL = {Atmospheric Chemistry and Physics},
VOLUME = {5},
YEAR = {2005},
NUMBER = {4},
PAGES = {1053--1123},
URL = {https://acp.copernicus.org/articles/5/1053/2005/},
DOI = {10.5194/acp-5-1053-2005}
}

@article{kraeutle2005new,
    author = {Kräutle, S. and Knabner, P.},
    title = {A new numerical reduction scheme for fully coupled multicomponent transport-reaction problems in porous media},
    journal = {Water Resources Research},
    volume = {41},
    number = {9},
    doi = {10.1029/2004WR003624},
    year = {2005}
}

@article{kraeutle2007reduction,
    author = {Kräutle, S. and Knabner, P.},
    title = {A reduction scheme for coupled multicomponent transport-reaction problems in porous media: {G}eneralization to problems with heterogeneous equilibrium reactions},
    journal = {Water Resources Research},
    volume = {43},
    number = {3},
    doi = {10.1029/2005WR004465},
    year = {2007}
}

@article{hoffmann2010parallel,
	author = {Hoffmann, Joachim and Kr{\"a}utle, Serge and Knabner, Peter},
	doi = {10.1007/s10596-009-9173-7},
	issn = {1573-1499},
	journal = {Computational Geosciences},
	number = {3},
	pages = {421--433},
	title = {A parallel global-implicit 2-{D} solver for reactive transport problems in porous media based on a reduction scheme and its application to the {MoMaS} benchmark problem},
	volume = {14},
	year = {2010},
}

@article{hoffmann2012general,
	author = {Hoffmann, Joachim and Kr{\"a}utle, Serge and Knabner, Peter},
	doi = {10.1007/s10596-012-9304-4},
	issn = {1573-1499},
	journal = {Computational Geosciences},
	number = {4},
	pages = {1081--1099},
	title = {A general reduction scheme for reactive transport in porous media},
	volume = {16},
	year = {2012},
}

@phdthesis{hoffmann2010monograph,
  author = {Hoffmann, J.},
  title  = {Reactive Transport and Mineral Dissolution/Precipitation in Porous Media:Efficient Solution Algorithms, Benchmark Computations and Existence of Global Solutions},
  school = {Friedrich-Alexander-Universit{\"a}t Erlangen-N{\"u}rnberg},
  year   = {2010},
  urn    = {urn:nbn:de:bvb:29-opus-17735}
}

@Article{petri_sosaa,
AUTHOR = {Clusius, P. and Baykara, M. and Xavier, C. and Zhou, P. and Tyree, J. and Foreback, B. and \"Aij\"al\"a, M. and Graeffe, F. and Pet\"aj\"a, T. and Kulmala, M. and Paasonen, P. and Palmer, P. I. and Boy, M.},
TITLE = {Modelling the impact of anthropogenic aerosols on CCN concentrations over a rural boreal forest environment},
JOURNAL = {Atmospheric Chemistry and Physics},
VOLUME = {26},
YEAR = {2026},
NUMBER = {3},
PAGES = {1967--1992},
URL = {https://acp.copernicus.org/articles/26/1967/2026/},
DOI = {10.5194/acp-26-1967-2026}
}

@Article{Rissanen2019MION,
AUTHOR = {Rissanen, M. P. and Mikkil\"a, J. and Iyer, S. and Hakala, J.},
TITLE = {Multi-scheme chemical ionization inlet (MION) for fast switching of reagent
ion chemistry in atmospheric pressure chemical ionization mass spectrometry
(CIMS) applications},
JOURNAL = {Atmospheric Measurement Techniques},
VOLUME = {12},
YEAR = {2019},
NUMBER = {12},
PAGES = {6635--6646},
URL = {https://amt.copernicus.org/articles/12/6635/2019/},
DOI = {10.5194/amt-12-6635-2019}
}

@Article{Yuan2016ToFCIMS,
AUTHOR = {Yuan, B. and Koss, A. and Warneke, C. and Gilman, J. B. and Lerner, B. M. and Stark, H. and de Gouw, J. A.},
TITLE = {A high-resolution time-of-flight chemical ionization mass spectrometer
utilizing hydronium ions (H$_{3}$O$^{+}$ ToF-CIMS) for measurements of
volatile organic compounds in the atmosphere},
JOURNAL = {Atmospheric Measurement Techniques},
VOLUME = {9},
YEAR = {2016},
NUMBER = {6},
PAGES = {2735--2752},
URL = {https://amt.copernicus.org/articles/9/2735/2016/},
DOI = {10.5194/amt-9-2735-2016}
}

@Article{Song2024ICIMS,
AUTHOR = {Song, M. and He, S. and Li, X. and Liu, Y. and Lou, S. and Lu, S. and Zeng, L. and Zhang, Y.},
TITLE = {Optimizing the iodide-adduct chemical ionization mass spectrometry (CIMS) quantitative method for toluene oxidation intermediates: experimental insights into functional-group differences},
JOURNAL = {Atmospheric Measurement Techniques},
VOLUME = {17},
YEAR = {2024},
NUMBER = {17},
PAGES = {5113--5127},
URL = {https://amt.copernicus.org/articles/17/5113/2024/},
DOI = {10.5194/amt-17-5113-2024}
}

@inbook{ashu2026attention,
author = {Ashu, Valery and Liu, Zhi-Song and Ashu, Taiwo and Rupp, Andreas and Haario, Heikki},
year = {2026},
month = {06},
pages = {390-407},
title = {Attention-Enhanced CNN Surrogate for Inverse Parameter Estimation in Cellular Automata},
booktitle = {Advances in Evolutionary and Deterministic Methods for Design, Optimization and Control},
series    = {Computational Methods in Applied Sciences},
volume    = {60},
isbn = {978-3-032-21892-6},
doi = {10.1007/978-3-032-21893-3_22},
Publisher = {Springer}
}


%% file: nn.bib
@article{peng2025spin,
  title={SPIN-ODE: Stiff Physics-Informed Neural ODE for Chemical Reaction Rate Estimation},
  author={Peng, Wenqing and Liu, Zhi-Song and Boy, Michael},
  journal={arXiv preprint arXiv:2505.05625},
  year={2025}
}
